%% file: optimal_decision_trees.tex
\documentclass[twoside,11pt]{article}

\usepackage[abbrvbib, preprint]{jmlr2e}

\usepackage[lined, ruled, linesnumbered, noend, scleft, nofillcomment]{algorithm2e}
\usepackage{bm}
\usepackage{wrapfig}

\usepackage{subfig}
\usepackage{graphicx}

\usepackage[T1]{fontenc}
\usepackage{pgfplots}

\pgfplotsset{
	compat=newest,
	xlabel near ticks,
	ylabel near ticks
}

\usepackage{xcolor}

\SetCommentSty{mycommfont}

\newcommand{\ignore}[1]{}

\newtheorem{property}{Property}
\newtheorem{observation}{Observation}

\usepackage{lastpage}
\jmlrheading{22}{2021}{1-\pageref{LastPage}}{5/20; Revised
10/21}{12/21}{20-520}{Emir Demirovi\'c, Anna Lukina, Emmanuel Hebrard, Jeffrey Chan, James Bailey, Christopher Leckie, Kotagiri Ramamohanarao, and Peter J. Stuckey}
\ShortHeadings{MurTree: Optimal Decision Trees via Dynamic Programming and Search}{Demirovi\'c, Lukina, Hebrard, Chan, Bailey, Leckie, Ramamohanarao, and Stuckey}

\firstpageno{1}

\begin{document}

\title{MurTree: Optimal Decision Trees via \\Dynamic Programming and Search}

\author{\name Emir Demirovi\'c \email e.demirovic@tudelft.nl \\
		\name Anna Lukina \email a.lukina@tudelft.nl \\
       \addr Delft University of Technology\\
       Delft, The Netherlands\\
	\AND
	\name Emmanuel Hebrard \email hebrard@laas.fr \\
       \addr LAAS CNRS\\
       Toulouse, France
       \AND
       \name Jeffrey Chan \email jeffrey.chan@rmit.edu.au \\
       \addr RMIT University\\
       Melbourne, Australia
	\AND
	\name James Bailey \email baileyj@unimelb.edu.au\\
	\name Christopher Leckie \email caleckie@unimelb.edu.au\\
	\name Kotagiri Ramamohanarao \email kotagiri@unimelb.edu.au\\
       \addr University of Melbourne\\
       Melbourne, Australia\\
       \AND
       \name Peter J. Stuckey \email peter.stuckey@monash.edu \\
       \addr Monash University and DATA61\\
       Melbourne, Australia
       }

\editor{Luc De Raedt}

\maketitle

\begin{abstract}
\footnote{The paper has been published in JMLR'22: https://jmlr.csail.mit.edu/beta/papers/v23/20-520.html.}Decision tree learning is a widely used approach in machine learning, favoured in applications that require concise and interpretable models. Heuristic methods are traditionally used to quickly produce models with reasonably high accuracy. A commonly criticised point, however, is that the resulting trees may not necessarily be the best representation of the data in terms of accuracy and size. In recent years, this motivated the development of optimal classification tree algorithms that globally optimise the decision tree in contrast to heuristic methods that perform a sequence of locally optimal decisions. We follow this line of work and provide a novel algorithm for learning optimal classification trees based on dynamic programming and search. Our algorithm supports constraints on the depth of the tree and number of nodes. The success of our approach is attributed to a series of specialised techniques that exploit properties unique to classification trees. Whereas algorithms for optimal classification trees have traditionally been plagued by high runtimes and limited scalability, we show in a detailed experimental study that our approach uses only a fraction of the time required by the state-of-the-art and can handle datasets with tens of thousands of instances, providing several orders of magnitude improvements and notably contributing towards the practical use of optimal decision trees.
\end{abstract}


\section{Introduction}

The combination of simplicity and effectiveness has popularised decision tree models in the machine learning community. A notable advantage of these models is their interpretability, in particular when the tree is of small size. Figure \ref{fig:decisionTree} shows an example of such a model, which may be easily understood even by non-experts.

Despite its clear structure, constructing a decision tree to represent the data is a challenging computational problem ($\mathbf{NP}$-hard). Traditionally, models are built using heuristic methods, such as CART (\cite{breiman1984classification}), which iteratively optimise a local objective function. While these techniques have shown to be immensely valuable due to their ability to provide high quality trees in low computational time, the resulting tree is not guaranteed to be globally optimal, i.e., it may not necessarily be the best representation of the data in terms of accuracy, size, or other considerations such as fairness. 

An alternative to heuristic approaches is to construct \emph{optimal decision trees}, i.e., the best possible decision tree according to a given metric. The idea of computing optimal decision trees dates back to approximately the 1970s when constructing optimal decision trees was proven to be $\mathbf{NP}$-hard by \cite{NPhardTrees}.\footnote{Their proof is for the problem of finding a perfect tree minimising the expected number of feature tests. However, it can easily be adapted to maximising the accuracy under a constraint on the maximum depth.} As emphasised by \cite{bertsimas2017optimal}, while optimal decision trees have always been desirable, the authors of the CART algorithm (\cite{breiman1984classification}) found that such trees were computationally infeasible at the time, and hence heuristics were the only option.

Optimal decision trees are enticing for several reasons. It has been observed that a more accurate representation of the data offers better generalisation on unseen data (\cite{bertsimas2017optimal,verwer2017learning,verwer2019learning}). This has been reiterated in our experiments as well. Globally optimal trees are particularly important in socially-sensitive contexts, where optimality plays an important role in ensuring \emph{fairness} (\cite{aghaei2019learning}). In some applications, the goal is to optimise the size of the decision tree representing a given controller to save memory for embedded devices (\cite{ashok2020dtcontrol}). Decision trees, in particular those of smaller size, are desirable for formal methods when verifying properties of trained controllers (\cite{bastani2018verifiable}), as opposed to more complex machine learning models. In recent years, there has been growing interest in \emph{explainable artificial intelligence}. The basic premise is that machine learning models, apart from high accuracy, must also be able to explain their decisions to a (non-expert) human. This is necessary to increase human trust and reliability of machine learning in complex scenarios that are conventionally handled by humans. Optimal decision trees of small size naturally fit within the scope of explainable AI, as their reduced size is more convenient for human interpretation.

Decision tree learning may be defined as a mathematical optimisation program: an objective function is posed together with a set of constraints that encode the decision tree structure. An advantage of optimal algorithms over heuristic approaches is that they adhere precisely to the given specification. This allows a clear analysis and assessment of the suitability of the particular mathematical formulation for a given application. In contrast, in heuristic methods there is a discrepancy between the target learning problem and the goals of the heuristic algorithm, i.e., the methods may not directly optimise the tree according to the globally defined objective, but rather locally optimise a sequence of subproblems with respect to a surrogate metric. This leads to situations where it may be difficult to make conclusive statements on the learning problem definition, as the heuristic approach may not faithfully follow the desired metrics. For example, a specification might be deemed suboptimal not due to a flaw in the definition, but rather because of the inability of the heuristic algorithm to optimise according to the specification. 

Despite the appeal of optimal algorithms for decision trees, heuristic methods are historically the dominant approach due to computational reasons. Indeed, heuristic methods offer scalable algorithms that produce results in the order of seconds. However, as both algorithmic techniques and hardware advanced, \emph{optimal} decision trees
have become within practical reach and attracted growing interest from the research community. In particular, there has been a surge of successful methods in the past few years. These approaches use generic optimisation methods, namely integer programming (\cite{bertsimas2017optimal,verwer2017learning,verwer2019learning,aghaei2019learning,MIPneurips2020}), constraint programming (\cite{verhaeghe2019learning}), and SAT (\cite{narodytska2018learning, avellanedaefficient, janota2020sat}), and algorithms tailored to the decision tree problem (\cite{nijssen2010optimal,dl8, hu2019optimal, dl85, sparseICML}). The methods DL8 (\cite{dl8,nijssen2010optimal}) and DL8.5 (\cite{dl85,aglin2020pydl8}) are of particular interest as they can be seen as a starting point for our work. The DL8.5 approach has been shown to be highly effective, outperforming other approaches when computing full binary decision trees on binary data, demonstrating the value of specialising methods to exploit specific decision tree properties over generic optimisation approaches.

\textbf{Our Contribution}. While previous works use highly related ideas, the presentation and terminology may differ substantially. In this work, we unify and generalise successful concepts from the literature by viewing the problem through the lens of a conventional algorithmic framework, namely dynamic programming and search. We introduce novel algorithmic techniques that reduce computation time by orders of magnitude when compared to the state-of-the-art. We conduct an experimental study on a wide range of benchmarks from the literature to show the effectiveness of our approach and its components, and reiterate that optimal decision trees lead to better generalisation in terms of out-of-sample accuracy. In more detail, the contributions are as follows:

\begin{itemize}
\item \emph{MurTree} (Section \ref{section:murtreeframework}), an algorithm for computing optimal classification trees. Given an input dataset and a set of predicates, it computes a decision tree that minimises the number of misclassifications using the given predicates. The algorithm allows constraints on the depth and the number of nodes of the decision tree. The method may be extended with additional functionality, such as multi-classification, regression, the sparse decision tree objective, lexicographical minimisation of misclassification and size, anytime behaviour, and nonlinear metrics, as discussed in Section \ref{section:extensions}.

\item A clear high-level view of the algorithm using conventional algorithmic principles, namely dynamic programming and search, that unifies and generalises some of the ideas from the literature (Section \ref{section:high-level}).

\item A specialised algorithm for computing the optimal classification tree of depth two, which serves as the backbone of our algorithm (Section \ref{section:specialisedAlgorithm}). It uses a frequency counting method to avoid explicitly referring to the dataset when constructing the tree, substantially reducing the runtime of computing optimal trees. The technique is further augmented with an incremental technique that takes into account previous computations, providing orders of magnitude speed-ups. Counting and incremental construction ideas have been previously used in classical algorithms, such as counting sort, and in the frequent itemset mining community, e.g., \cite{diffsets}. We exploit such ideas  in the context of decision trees.

\item A novel similarity-based mechanism for computing a lower bound on the number of misclassifications. The bound is effective in determining that portions of the search space cannot contain better decision trees than currently found during the search, which allows the algorithm to prune parts of the search space without needing further inspection, providing additional speed-ups. The bound is derived by examining previously computed subtrees and computing a bound on the number of misclassifications that must hold in the new search space (Section \ref{section:similarityBounding}).

\item Several extensions to DL8.5 (\cite{dl85}), namely we incorporate the constraint on the number of nodes, extend the caching technique to take into account constraints on both the depth and number of nodes and provide a novel implementation of two existing caching schemes (Section \ref{section:caching}), describe an incremental solving option to allow reusing computation when solving a series of increasingly large decision trees (Section \ref{section:incrementalsolving}), which is useful in hyper-parameter tuning, for example, refine the lower bounding technique on the number of misclassifications from DL8.5 (Section \ref{section:LBstore}), and discuss a dynamic node exploration strategy (Section \ref{section:nodeselection}) that leads to consistent improvements over a conventional post-order search.

\item A detailed experimental study to analyse the effectiveness of our individual techniques and scalability of our approach, evaluate our approach with respect to the state-of-the-art optimal classification tree algorithms, and compare against heuristic decision tree and random forest algorithms on out-of-sample accuracy (Section \ref{section:experiments}). The experimental results show that our approach provides generalisable trees and exhibits speed-ups of (several) orders of magnitude when compared to the state-of-the-art.

\end{itemize}

The rest of the paper is organised as follows. In the next section, we introduce the notations and definitions used throughout the paper. In Section \ref{section:literature}, we review the state-of-the-art for optimal decision trees. Our main contribution is given in Section \ref{section:murtreeframework}, where we describe our \emph{MurTree} algorithm. In Section \ref{section:experiments}, we conduct a series of empirical evaluations of our approach and conclude in Section \ref{section:conclusion}.  

\section{Preliminaries}

A \emph{feature} is a variable that encodes information about an object. 
We speak of \emph{binary}, \emph{categorical}, and \emph{continuous} features depending on their domain, i.e., binary, discrete, and continuous domains.
A \emph{feature vector} is a vector of features. An \emph{instance} is a pair that consists of a feature vector and a value representing the \emph{class}. A class can take continuous or discrete values. In future text, we assume the class is a discrete value, i.e. we consider \emph{classification} tasks. A \emph{dataset}, or simply \emph{data}, is a set of instances. While features within a vector may have different domains, the \emph{i-th} feature of each feature vector of the dataset shares the same domain. The assumption is that the features describe certain characteristics about the objects, and the \emph{i-th} feature of each feature vector refers to the same characteristic.

\begin{figure}
\center
\includegraphics[width=0.7\linewidth]{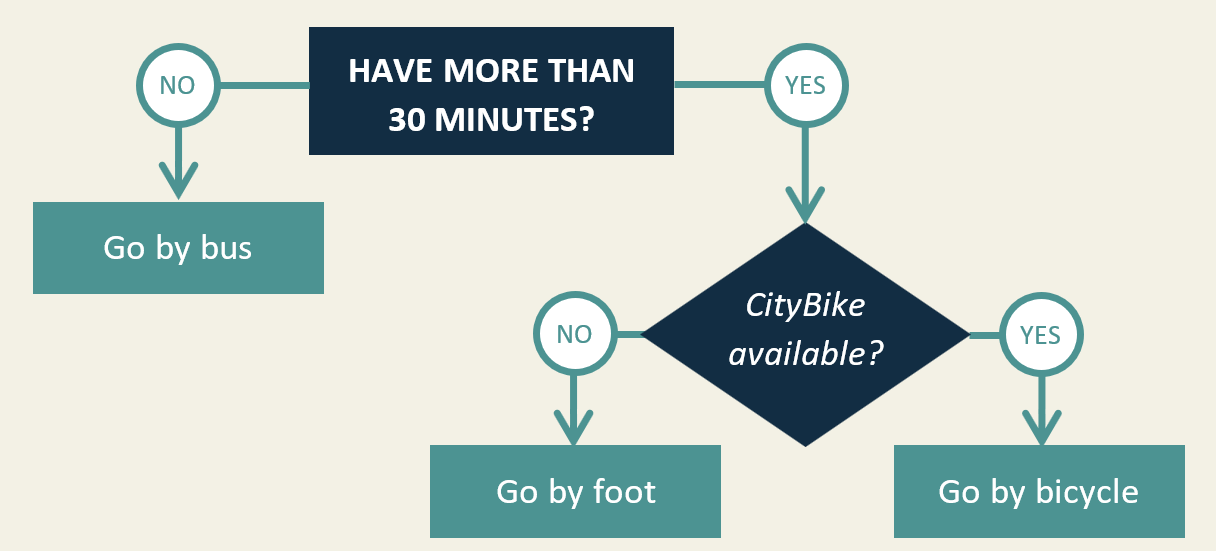}
\caption{A Decision Tree for Commuting to Work}
\label{fig:decisionTree}
\end{figure}

A \emph{decision tree} is a machine learning model that takes the form of a tree (see Figure \ref{fig:decisionTree}). We consider binary trees, i.e., trees that contain nodes with at most two children. We call leaf and non-leaf nodes \emph{classification} and \emph{predicate} nodes, respectively. Each predicate node is assigned a predicate that maps feature vectors to a Boolean value, e.g., \emph{``CityBike available?''} is a predicate with a clear yes/no answer. The left and right edges of a predicate node are associated with the values zero and one, respectively. Each classification node is assigned a fixed class. We note that other variations of decision trees are possible, e.g., more than two children, but these are not considered in this work. 

A decision tree may be viewed as a function that performs classification according to the following recursive procedure. Given a feature vector, it starts by considering the root node. If the considered node is a classification node, its class determines the class of the feature vector and the procedure terminates. Otherwise, the node is a predicate node, and the left child node will be considered next if the predicate of the node evaluates to zero, and otherwise the right child node is selected. The process recurses until a class is determined. The \emph{misclassification score} of a decision tree on data is the number of instances for which the classification produces an incorrect class considering the data as ground truth.

In practice, the predicates take a special form. For \emph{single-variate} or \emph{axis-aligned} decision trees, which are the focus of this work, predicates only consider a single feature and typically test whether it exceeds a threshold value. For example, the predicate in Fig. \ref{fig:decisionTree} \emph{``Have more than 30 minutes?''} considers a single feature representing the available time in minutes and tests whether it exceeds the value thirty. We refer to these nodes as \emph{feature nodes}, as the predicate depends solely on one feature. Predicates are chosen based on the dataset. Generalisations of decision trees are straight-forward: \emph{multi-variate} versions use predicates that operate on more than one feature, and predicates can be substituted by functions whose co-domains are of size $n$, in which case the decision tree is an $n$-ary tree with an analogous definition. These generalisations are not considered in this work.


The \emph{depth} of a decision tree is the maximum number of feature nodes any instance may encounter during classification. The \emph{size} of a decision tree is the number of feature nodes. For example, the decision tree in Fig. \ref{fig:decisionTree} has the depth and size equal to two. It follows that the maximum size of a decision tree with depth $d$ is $2^d-1$. An alternative size definition may consider the total number of nodes in the tree. Note that these definitions are equivalent, as a tree with $n$ predicate nodes has $n+1$ classification nodes. 

The process of \emph{decision tree learning} seeks to compute a decision tree that minimises a target metric under a set of constraints. We are primarily concerned with minimising the misclassification score given a maximum depth and a maximum number of feature nodes. Other metrics and constraints may also be considered (see Section \ref{section:extensions}). Given the misclassification score, classification nodes will be assigned the class that minimises the misclassifications on the given dataset. This corresponds to computing the instances that reach the classification node during classification and selecting the class of the node according to the majority class.


Following previous optimal decision tree works (\cite{dl8,nijssen2010optimal,narodytska2018learning,verhaeghe2019learning,dl85,hu2019optimal,sparseICML}), we consider the setting where all features are binary. Datasets with continuous and/or categorical features are assumed to be \emph{binarised} as a preprocessing step. This corresponds to selecting the set of predicates upfront rather than during algorithm execution as is standard in heuristic algorithms. 

We use special notation for \emph{binary datasets}, where the domain of
features is Boolean, i.e., $\{0, 1\}$. 
Given a feature vector $\mathbf{fv}$, we
write $f_i \in \mathbf{fv}$ and $\overline{f_i} \in \mathbf{fv}$ if the
\emph{i-th} feature has value one and zero, respectively. 
If $f_i \in \mathbf{fv}$, we say the \emph{i-th} feature is \emph{present} in the feature vector $\mathbf{fv}$ (a \emph{positive feature}), otherwise it is not present (a \emph{negative} feature). 
We consider only predicates that test the presence of a feature, i.e., the predicates $P_i(\mathbf{fv})$ and $\overline{P_i}(\mathbf{fv})$ evaluate to \emph{one} if $f_i \in \mathbf{fv}$ and $\overline{f_i} \in \mathbf{fv}$, respectively, and evaluate to \emph{zero} otherwise. Given this special form, we simply write $f_i$ or $\overline{f_i}$ for the predicates instead of $P_i$ and $\overline{P_i}$.
The binary dataset $\mathcal{D}$ is partitioned into a positive and negative class of instances based on the classes, i.e., $\mathcal{D} = \mathcal{D^+} \cup\mathcal{D^-}$. 
We consider the partitions as sets of feature vectors since their class is clear from context, and write $\mathcal{D}(f)$ as the set of instances from $\mathcal{D}$ that contain feature $f$, and analogously for multiple features, e.g., $\mathcal{D}(f_1, f_2)$ are the set of instances that contain both $f_1$ and $f_2$.

\section{Literature Review}
\label{section:literature}

Historically, the most popular techniques for decision tree learning were based on heuristics due to their effectiveness and scalability. Examples of these algorithms include CART, originally proposed by \cite{breiman1984classification}, and C4.5 by \cite{c4-5}. These algorithms start with a single node, and iteratively expand the tree based on metrics such as information gain and Gini coefficients, possibly post-processing the obtained decision trees to prune branches in an effort to reduce overfitting. While there is a vast literature on (heuristic) algorithms for decision trees, in this work, we are primarily concerned with \emph{optimal single-variate decision trees}, and hence direct further discussion to such settings.

\cite{bertsimas2007classification} presented a mixed-integer programming approach for optimal decision tree	s that worked well on smaller datasets. Mixed-integer programming formulations with better performance were given by \cite{bertsimas2017optimal} and \cite{verwer2017learning}. These methods encode the optimal decision tree by fixing the tree depth in advance, creating variables to represent the predicates for each node, and adding constraints to enforce the decision tree structure. These approaches were later improved by \emph{BinOPT} (\cite{verwer2019learning}), a \emph{binary linear programming} formulation, that took advantage of binarising data to reduce the number of variables and constraints required to encode the problem.
\cite{aghaei2019learning} used a mixed-integer programming formulation for optimal decision trees that supported \emph{fairness} metrics. The authors argued that using machine learning in socially sensitive contexts may perpetuate discrimination if no special measures are taken into account. In this instance, optimal decision trees provide the best tree that balanced accuracy and fairness, albeit with a high computational time when compared to specialised heuristic methods (\cite{kamiran2010discrimination}).  Recently, \cite{MIPneurips2020} proposed a novel mixed-integer programming formulation based on support vector machines and a cutting plane technique for optimal multi-variate decision trees, and a flow-based encoding has been developed \cite{aghaei2020learning}. An advantage of declarative approaches is that adding additional constraints may be straight-forward, however scalability may be an issue when compared to specialised approaches when considering single-variate trees, e.g., DL8.5 (see below) or our method. For more information regarding decision tree optimisation using mathematical programming, we refer the readers to the survey by \cite{surveyMathematicalTrees}.


Encodings of decision trees using propositional logic (SAT) and constraint programming have been initially devised by \cite{firstcp}. Recently,
an improved SAT model has been proposed by~\cite{narodytska2018learning}, after which several other SAT-related works have been published (\cite{avellanedaefficient,janota2020sat,LNSSAT}). This line of work deviates from conventional machine learning approaches, as the aim is to construct the smallest tree that \emph{perfectly} describes the given dataset, i.e., leads to zero misclassifications on the training data, although they can be adapted to the accuracy criterion via maximum satisfiability (MaxSAT)~\cite{sataccuracy}.
To circumvent scalability issues, the methods perform subsampling of the data, incrementally construct the encoding, and/or focus on improving a subtree obtained using a heuristic algorithm.


\cite{dl8,nijssen2010optimal} introduced a framework named \emph{DL8} for optimal decision trees that could support a wide range of constraints. They observed that the left and right subtree of a given node can be optimised independently, introduced a caching technique to save subtrees computed during the algorithm in order to reuse them at a later stage, and combined these with ideas from the pattern mining literature to compute optimal decision trees. DL8 laid a foundation for optimal decision tree algorithms that follow.

\cite{verhaeghe2019learning} approached the optimal classification tree problem by minimising the misclassifications using constraint programming. The independence of the left and right subtrees from \cite{dl8,nijssen2010optimal} was captured in an AND-OR search framework. Upper bounding on the number of misclassifications was used to prune parts of the search space and their algorithm incorporated an itemset mining technique to speed-up the computation of instances per node and used a caching technique similar to DL8 (\cite{dl8,nijssen2010optimal}).

\cite{hu2019optimal} presented an algorithm that computes the optimal decision tree by considering a balance between misclassifications and number of nodes. They apply exhaustive search, caching, and lower bounding of the misclassifications based on the cost of adding a new node to the decision tree. The method was improved and extended by \cite{sparseICML}, providing good performance if the sparsity coefficient, which controls the balance between accuracy and number of nodes, is sufficiently high.

\cite{dl85} developed \emph{DL8.5} by combining and refining the ideas from \emph{DL8} and the constraint programming approach. Their main addition was an upper bounding technique, which limited the upper misclassification value of a child node once the optimal subtree was computed for its sibling, and a lower bounding technique, where the algorithm stored information not only about computed optimal subtrees but also pruned subtrees to provide a lower bound on the misclassifications of a subtree. This led to an algorithm that outperformed previous approaches by a notable margin when optimising the misclassification score under a depth constraint. The method was recently released as a Python library with further improvements based on sparse bitvectors (\cite{aglin2020pydl8}).

Exploiting properties specific to the decision tree learning problem proved to be valuable in improving algorithmic performance in previous work. In particular, search and pruning techniques, caching computation for later reuse, and the techniques that  take advantage of the decision tree structure all lead to notable gains in performance. These are the main reasons for the success of specialised methods over generic frameworks, such as integer programming and SAT. As there is a significant overlap of ideas and techniques used in related work, we discuss these in more detail in Section \ref{section:high-level} when presenting the high-level view of our algorithm.

The above discussion was mainly concerned with single-variate optimal decision tree algorithms, which are the focus of this work. Other related work includes heuristic methods for multi-variate trees (\cite{yang2019weighted}), theoretical analysis of heuristic methods (\cite{blanc2020provable}), a fine-grained computational complexity study (\cite{parametrizedTrees}), neural networks for decision trees (\cite{kontschieder2015deep,tanno2019adaptive}), randomised trees (\cite{blanquero2020sparsity}), end-to-end learning of decision trees (\cite{hehn2019end,elmachtoub2020decision}), and dynamic programming methods to construct decision trees from random forests (\cite{icmlForest}). For more refereces, we refer the readers to a curated list of decision tree papers by Benedek Rozemberczki: \url{https://github.com/benedekrozemberczki/awesome-decision-tree-papers}.

\section{MurTree: Our Algorithm for Optimal Classification Trees}
\label{section:murtreeframework}

Our algorithm computes optimal classification trees by exhaustive search. The search space is exponentially large, but special measures are taken to efficiently iterate through all trees, exploit the overlap between trees, and avoid computing suboptimal decision trees. We give the main idea of the algorithm, then provide the full pseudocode, and follow up with individual subsections where we present each individual technique in greater detail. 

The following text focusses on optimal classification trees that minimise the number of misclassified instances for binary datasets and binary classification given constraints on the depth and number of nodes. This serves as the core part of our algorithm. Further extensions are discussed in Section \ref{section:extensions}, which includes multi-classification, regression, optimising the sparse objective, lexicographically minimising the misclassification score and the number of nodes, anytime behaviour, and optimising nonlinear metrics.

\subsection{High-Level Idea}
\label{section:high-level}

We note two important properties of decision trees:

\begin{property}(Independence)
\label{property:independence}
Given a dataset $\mathcal{D}$, a feature node partitions the dataset $\mathcal{D}$ into its left and right subtree, such that $\mathcal{D}_{left} \cap \mathcal{D}_{right} = \emptyset$ and $\mathcal{D} = \mathcal{D}_{left} \cup \mathcal{D}_{right}$.
\end{property}

\begin{property}(Overlap)
\label{property:overlap}
Given a classification node, a set of features encountered on the path from the root node to the classification node, and an instance, the order in which the features are used to evaluate the instance does not change the classification result. 
\end{property}

Both properties follow directly from the definition of decision trees and are emphasised as they play a major role in designing decision tree algorithms. Property \ref{property:independence} allows computing the misclassification score of the tree as the sum of the misclassification scores of its left and right subtree. As will be discussed, this is important as once a feature node has been selected, the left and right subtrees can be optimised independently of each other. Property \ref{property:overlap} shows there is an overlap between decision trees that share the same features, which is taken advantage of by caching techniques. For example, once the optimal tree has been computed for the dataset $\mathcal{D}(f_1, f_2)$, the resulting tree is stored in the cache and reused when the dataset $\mathcal{D}(f_2, f_1)$ is encountered (see Section \ref{section:caching} for more details on caching), since both $\mathcal{D}(f_1, f_2)$ and $\mathcal{D}(f_2, f_1)$ represent exactly the same subproblem. 

The dynamic programming formulation of optimal classification trees given in
Eq.~\ref{DPformulation} provides a high-level summary of our algorithm. The
input parameters consist of a binary dataset $\mathcal{D}$ with features
$\mathcal{F}$, an upper bound on depth $d$, and an upper bound on the 
number of feature nodes $n$. The output is the minimum number of misclassifications possible on the data given the input decision tree characteristics. 


\begin{equation}
\label{DPformulation}
T(\mathcal{D}, d, n) = 
\left\{ \begin{array}{ll}
T(\mathcal{D}, d, 2^d-1) & n > 2^d-1\\
T(\mathcal{D}, n, n) & d > n\\
\min\{|\mathcal{D}^+|, |\mathcal{D}^-|\} &n = 0 \lor d = 0\\
 \min\{T(\mathcal{D}(\overline{f}), d-1, n - i - 1)   & n > 0 \land d > 0\\
~~~~~+ T(\mathcal{D}(f), d-1, i) : f \in \mathcal{F}, i \in [0, n-1]\}
\end{array}
\right.
\end{equation}

The first and second case in Eq.~\ref{DPformulation} place a natural limit on the number of feature nodes and depth to avoid redundancy. The third case captures the situation where the node must be declared as a classification node, i.e., the node is labelled according to the majority class. The fourth case states that computing the optimal misclassification score amounts to examining all possible feature splits and ways to distribute the feature node count to the left and right children of the root node. For each combination of a selected feature and node count distribution to its children, the optimal misclassification is computed recursively as the sum of the optimal misclassifications of its children. The formulation is exponential in the depth, feature node limit, and number of features, but with special care, as presented in the subsequent sections, it is possible to compute practically relevant optimal classification trees within a  reasonable time.

Eq.~\ref{DPformulation} serves as the core foundation of our algorithm. In contrast to previous algorithms, we take advantage of the structure of decision trees to allow imposing a limit on the number of nodes as presented in Eq.~\ref{DPformulation}. For example, previous methods either set the number of nodes to the maximum value given the depth (\cite{dl85,avellanedaefficient}), do not directly limit the number of nodes but instead penalise the objective function for each node in the tree (\cite{hu2019optimal,sparseICML}), or allow constraints on the number of nodes but do not make use of decision tree properties (\cite{bertsimas2017optimal,narodytska2018learning,verwer2017learning,verwer2019learning}). The last point is particularly important as the ability to exploit optimal decision tree properties has proven to be essential in achieving the best performance. 

Different forms of Eq. \ref{DPformulation} were used in some
previous work under different terminology. The AND-OR search method
(\cite{verhaeghe2019learning}), pattern mining approach (\cite{dl8,nijssen2010optimal,dl85}),
and the search by \cite{hu2019optimal} and \cite{sparseICML} use the independence property of the
left and right subtree (Property \ref{property:independence}). Those
approaches save computed optimal subtrees (Property \ref{property:overlap}),
which corresponds to \emph{memoisation} as an integral part of dynamic
programming (Section \ref{section:caching}). The DL8 papers (\cite{dl8,nijssen2010optimal}) introduced a general framework with a variety of constraints which includes, amongst others, constraints on the depth, node count, and fairness. Framing the
problem as a dynamic program dates from the 1970s 
(e.g., \cite{garey1972optimal}), 
but the description in works afterwards deviated as new techniques were introduced. Our contribution is presenting the problem using conventional dynamic programming notation and algorithms that respect constraints on the depth of the tree and the number of nodes.

A key component of our algorithm is a specialised method for computing decision trees of depth at most two. It takes advantage of the specific decision tree structure by performing a precomputation on the data, which allows it to compute the optimal decision tree without explicitly referring to the data. This offers a significantly lower computational complexity compared to the generic case of Eq. \ref{DPformulation}, but is applicable in practice only to decision trees of depth two. Thus, rather than following Eq. \ref{DPformulation} until the base case, we stop the recursion once a tree of depth two is required and invoke the specialised method.

A defining characteristic of search algorithms are pruning techniques, which detect areas of the search that may be discarded without losing optimality. In the case of decision trees, subtrees may be pruned based on the lower or upper bound\footnote{Note that the term `upper bound' is to \emph{not} meant to be interpreted as an upper bound to the global problem in the strict mathematical sense, but rather as a value that when exceeded leads to trees that have more misclassifications than the currently best known tree during the execution of the algorithm.} of the number of misclassifications of the given subtrees. If the lower bound shows that the misclassifications of a currently considered subtree will result in a value greater than the set upper bound, the subtree can be pruned, effectively reducing the search space. Note that the upper bound is always set in a way to exclude trees that have a higher misclassification score than the best tree found so far. The challenge when designing bounding techniques is to find the correct balance between pruning power and the computational time required by the technique. 

We introduce a novel similarity-based lower bounding technique (Section \ref{section:similarityBounding}) that derives a bound based on the similarity of the previously considered subtrees. We use our lower bounding method in combination with the previous lower bounding approach introduced in DL8.5 (\cite{dl85}), which we describe in the following text. Given a parent node, once the optimal subtree is computed for one of the children, an upper bound can be posed on the other child subtree based on the best decision tree known for the parent node and the number of misclassifications of the optimal child subtree. If a subtree fails to produce a solution within the posed upper bound, the upper bound is effectively a lower bound that can be used once the same subtree is encountered again in the search. Our algorithm uses a refinement of the described lower bound, which additionally takes into account all lower bounds of the children of the parent node (Section \ref{section:LBstore}).

The next subsection describes our techniques in more detail.

\subsection{Main Algorithm Description}
\label{section:mainloop}

Algorithm \ref{algo_main} summarises our algorithm. It can be seen as an instantiation of Eq. \ref{DPformulation} with additional techniques to speed-up the computation. In further text, we use the convention that \emph{infeasible trees} (denoted with $\emptyset$) have an infinite misclassification score.

The algorithm takes as input a dataset $\mathcal{D}$ consisting of positive $\mathcal{D}^+$ and negative $\mathcal{D}^-$ instances, branch information (initially empty, see below for details), the depth and the number of feature nodes, and an upper bound that represents a limit on the number of misclassifications before the tree is deemed \emph{infeasible}, i.e., not of interest for example since a better tree is known. The output is an optimal classification tree respecting the input constraints on the depth, size, and upper bound, or a flag indicating that no such tree exists, i.e., the problem is \emph{infeasible}. The latter occurs as a result of recursive calls (see further), which pose an upper bound that is necessary to ensure the decision tree has a lower misclassification value than the best tree found so far. The upper bound is initially set to the misclassification score of a single classification node for the data and is updated throughout the execution. We note that a tighter upper bound could be computed by using a heuristic method at the start. The algorithm proceeds as follows.

(Alg. \ref{algo_main}: lines \ref{alg1:line:UBtestSTART}-\ref{alg1:line:UBtestEND}) If the upper bound is negative, the algorithm reports infeasibility. Negative bounds may be a result of the calls in the general case algorithm (Alg. \ref{main_algo_general_case} and \ref{main_algo_general_case_helper}).

(Alg. \ref{algo_main}: lines \ref{alg1:line:basecaseSTART}-\ref{alg1:line:basecaseEND}) If no feature nodes are allowed, the algorithm returns a classification node or reports infeasibility in case the classification node misclassification exceeds the upper bound. The method \emph{LeafMisclassification} computes the misclassification score of a classification node given a dataset $\mathcal{D}$ as $\min\{|\mathcal{D}^-|, |\mathcal{D}^+|\}$, and the method \emph{ClassificationNode} returns a tree consisting of a single classification node that minimises the misclassification score on the dataset $\mathcal{D}$, i.e., it assigns the majority class as its label.

(Alg. \ref{algo_main}: lines \ref{alg1:line:cacheTestSTART}-\ref{alg1:line:cacheTestEND}) After basic tests, the cache is queried to check whether the optimal subtree has already been computed as part of a previous recursive call. If the optimal subtree is present in the cache, it is returned if the optimal subtree meets the upper bound constraint, otherwise infeasibility is reported. Caching subtrees for trees where the depth is constrained dates from DL8 (\cite{dl8,nijssen2010optimal}). In our work, the algorithm additionally caches with respect to the depth and number of node constraints. 

(Alg. \ref{algo_main}: lines \ref{alg1:line:cacheUpdateSTART}-\ref{alg1:line:cacheUpdateEND}) Assuming that the optimal subtree is not in the cache, the cache is updated using our similarity-based lower bound (Section \ref{section:similarityBounding}). Naturally, the new lower bound will replace the old cached bound only if it is of greater value. In case the lower bounding procedure happens to recover an optimal solution for the subtree, it is returned or infeasibility is reported if it exceeds the upper bound (see Section \ref{section:similarityBounding} for details).

(Alg. \ref{algo_main}: lines \ref{alg1:line:LBPruneSTART}-\ref{alg1:line:LBPruneEND}) Afterwards, the algorithm attempts to prune based on the (possibly updated) lower bound stored in the cache, or return a classification node if the lower bound matches the misclassification score of the classification node. 

(Alg. \ref{algo_main}: lines \ref{alg1:line:specialisedAlgSTART}-\ref{alg1:line:specialisedAlgEND}) After all simpler operations have been performed, the algorithm tests if the subproblem is a tree of depth at most two. A key aspect of our algorithm is that trees of depth at most two are computed using a specialised procedure (Section \ref{section:specialisedAlgorithm}). Should this be the case, the specialised algorithm is used to solve the subtree and store in the cache the solutions using one, two, and three feature nodes, regardless of the input requirements (number of nodes and upper bound). This is done since running the specialised algorithm produces the mentioned solutions as part of its procedure and it may be beneficial to store all the results. Once the computation is done, the algorithm returns the corresponding subtree given the input number of nodes if it is within the upper bound limit, otherwise reports infeasibility.

(Alg. \ref{algo_main}: line \ref{alg1:line:recursion}) Assuming neither of the above conditions took place, the algorithm reaches the general (fourth) case from Eq. \ref{DPformulation}, where the search space is exhaustively explored through a series of overlapping recursions. This is detailed in Alg. \ref{main_algo_general_case} and summarised below. 

\emph{Algorithm \ref{main_algo_general_case}}: General Case. (Alg. \ref{main_algo_general_case}: line \ref{alg2:line:split}) The algorithm considers each feature split. (Alg. \ref{main_algo_general_case}: line \ref{alg2:line:LBSTART}-\ref{alg2:line:LBEND}) If the current best solution is at the lower bound, the optimal decision tree has been found and no further splits need to be considered. (Alg. \ref{main_algo_general_case}: line \ref{alg2:line:discriSTART}-\ref{alg2:line:discriEND}) Feature splits that do not discriminate at least one instance are skipped since they add no value to the tree. This is summarised as follows.

\begin{definition}(Degenerate Decision Trees) 
A decision tree is degenerate if it contains at least one predicate node that does not discriminate a single training instance, i.e., the predicate returns the same truth value for each of its training instances.
\end{definition}

\begin{proposition}(Pruning Degenerate Trees) 
\label{degeneratePruning}
Given a degenerate decision tree with $n$ feature nodes and misclassification score $s$ on the training data, there exists at least one other decision tree with $n' < n$ feature nodes and misclassification score $s' \leq s$.
\end{proposition}


(Alg. \ref{main_algo_general_case}: line \ref{alg2:line:nodeBudgetSTART}-\ref{alg2:line:nodeBudgetEND}) Recall that the number of allowed feature nodes is given as input. Once a feature has been considered as the root node (line \ref{alg2:line:split}), the remaining node budget is split among its children. For each feature split, the algorithm considers all possible combinations of distributing the remaining node budget amongst its left and right subtrees. Note that when considering no other node limit other than the limit imposed by the depth of the tree, there is only one such combination, i.e., $n_{max} = n_{min}$ in Algorithm \ref{main_algo_general_case}. The algorithm invokes a subroutine given in Algorithm \ref{main_algo_general_case_helper} (see below), which computes the optimal tree given the tree configuration (the feature of the root and the number of features in children subtrees), or reports a \emph{local lower bound} on the solution (Section \ref{section:LBstore}).

(Alg. \ref{main_algo_general_case}: line \ref{alg2:line:cacheUpdatesSTART}-\ref{alg2:line:cacheUpdatesEND}) Throughout the algorithm the best tree found so far is recorded (line \ref{alg2:line:bestUpdate}). After exhausting all feature splits for the root node, if the best tree found is within the upper bound limit, the best tree is the optimal decision tree for the considered subproblem and it is stored in the cache. Otherwise, a lower bound is computed and stored in the cache. The fact that no tree was found within the upper bound limit implies a lower bound for the given subproblem is one greater than the input upper bound. The information is stored in the cache in case the bound is needed in one of the other recursive calls. This bound was introduced in DL8.5 \cite{dl85} and we provide a further refinement by taking into account the local bounds (see the refined bound in Section \ref{section:LBstore}). 

At the end of Algorithm \ref{main_algo_general_case}, the internal data structures related to our similarity lower bound are updated using the dataset $\mathcal{D}$ (see Section \ref{section:similarityBounding}) before returning the result of the algorithm, i.e., either the optimal decision tree or an infeasible tree indicating that no such tree exists within the input specification.

\emph{Algorithm \ref{main_algo_general_case_helper}}: Subroutine  to compute the optimal tree for a given tree configuration. For a chosen tree configuration (the feature of the root and the number of features in children subtrees), the algorithm determines the maximum depth of the left and right subtrees based on the second case of Eq. \ref{DPformulation}. It then considers which subtree to recurse on first. Previous work in DL8.5 fixed the order by exploring the left before the right subtree. In our algorithm, we introduce a dynamic strategy that prioritises the subtree with the greater misclassification score of its leaf node (Section \ref{section:nodeselection}). The intuition is that such a subtree is more likely to result in a tree with more misclassifications, and if one subtree has a high misclassification score it increases the likelihood of pruning the other sibling. For example, given a scenario where there are two children, one that will result a zero misclassification score and one that will result in an infeasible subtree, exploring the later first will remove the need to process the former, whereas processing the nodes in reverse order will require solving both subtrees instead of only one.


The algorithm then solves the subtrees in the chosen order. When computing the upper bound of the first subtree, the lower bound of the second is taken into account together with the global upper bound provided. If the first subtree is infeasible, a \emph{local lower bound} is returned using Eq. \ref{eq:localBound}. Otherwise, the second subtree is computed. If both the left and right subtree calls successfully terminated, the obtained decision tree is returned as the optimal tree. Otherwise, a local lower bound is returned. 

This concludes the main description of our algorithm. Before proceeding with detailing each component of our algorithm, we reiterate the differences between our approach and DL8.5 (\cite{dl85}) in light of the technical description given above.


\emph{Comparison with DL8.5 (\cite{dl85})}. Algorithm \ref{algo_main} shares a similar layout as in DL8.5, but there are notable differences that result in orders of magnitude speed-ups. The differences can be summarised as follows: 1) we allow constraining the number of feature nodes in addition to the depth, which is important in obtaining the smallest optimal decision, e.g., to improve interpretability, or when hyper-parameter tuning to learn trees that better generalise on unseen instances (Section \ref{section:outofsample}), 2) our specialised algorithm (Section \ref{section:specialisedAlgorithm}) is substantially more efficient at computing trees with depth two when compared to the general algorithm in Algorithm \ref{algo_main} or DL8.5, 3) we propose a new lower bound based on the similarity with previously computed subtrees to further prune the search space (Section \ref{section:similarityBounding}), refine the previous lower bound (Section \ref{section:LBstore}), and consider the lower bound when imposing the upper bound on the subtrees, 4) our cache policy (Section \ref{section:caching}) is extended to support the number of feature nodes constraint and allows for \emph{incremental solving}, allowing reusing computation when solving trees with new depth or number of nodes, e.g., during hyper-parameter tuning, 5) we dynamically determine which subtree to explore first based on pruning potential (Section \ref{section:nodeselection}) rather than use a static strategy, 6) we discuss our novel implementation of two caching strategies (based on branches and datasets) that leads to a light-weight cache (Section \ref{section:caching}), allowing us to take advantage of more advanced hashing on the dataset rather than on branches as done in DL8.5, and 7) we propose a number of extensions (see Section \ref{section:extensions}). 

\input{algo_main}

\input{algo_main_general_case}

\input{algo_main_general_case_2}




	
\subsection{Specialised Algorithm for Trees of Depth Two}
\label{section:specialisedAlgorithm}

An essential part of our algorithm is a specialised method for computing optimal decision trees of depth two. The procedure is repeatedly called in our algorithm, i.e., each time a tree of at most depth two needs to be optimally solved. In the following, we present an algorithm that achieves lower complexity than the general algorithm (Eq. \ref{DPformulation} and Prop. \ref{prop:generaldepth2}) when considering trees with depth two.

Prior to presenting our specialised algorithm, we discuss the complexity of computing decision trees of depth two using Eq. \ref{DPformulation} as the baseline.

\begin{proposition}
\label{prop:generaldepth2}
Computing the optimal classification tree of depth two using Eq. \ref{DPformulation} can be done in $\mathcal{O}(|\mathcal{D}| \cdot |\mathcal{F}|^2)$ time. 
\end{proposition}

Assume that splitting the data based on a feature node is done in $\mathcal{O}(|\mathcal{D}|)$ time. Eq. \ref{DPformulation} considers $|\mathcal{F}|$ splits for the root and for each feature performs $2 \cdot |\mathcal{F}|$ splits for its children. This results in $2 \cdot |\mathcal{F}|^2$ splits and an overall runtime of $\mathcal{O}(|\mathcal{D}| \cdot |\mathcal{F}|^2)$, proving Proposition \ref{prop:generaldepth2}. In practice, partitioning the dataset based on a feature can be sped-up using bitvector operations and caching subproblems (\cite{dl85,verhaeghe2019learning,hu2019optimal}), but the complexity remains as this only impacts the hidden constant in the big-O.

In the following, we present an algorithm with lower complexity and additional practical improvements which, when combined, reduce the runtime of computing the optimal classification tree of depth two by orders of magnitudes.

Algorithm \ref{algorithm:specialisedDT} provides a summary. The input is a dataset $\mathcal{D}$ and the output is the optimal classification tree of depth two with three feature nodes that minimises the number of misclassified instances. 

The specialised procedure computes the optimal decision tree in two phases. In the first step, it computes frequency counts for each pair of features, i.e., the number of instances in which both features are present. In the second step, it exploits the frequency counts to efficiently enumerate decision trees without needing to explicitly refer to the data. This provides a substantial speed-up compared to iterating through features and splitting data as given in the dynamic programming formulation (Eq.~\ref{DPformulation}) for decision trees of depth two. We now discuss each phase in more detail and present a technique to incrementally compute the frequency counts. We note that related counting and incremental computation ideas have been used in classical algorithms, such as counting sort, and frequent itemset mining methods, e.g., \cite{diffsets}.

\input{algo_specialised}

\subsubsection{Phase One: Frequency counting (Algorithm \ref{algorithm:specialisedDT}, Lines \ref{specialisedtree:PhaseOneStart}-\ref{specialisedtree:PhaseOneEnd})}

Let $FQ^{+}(f_i)$ and $FQ^{+}(f_i, f_j)$ denote the frequency counts  in the positive instances for a single feature and a pair of features, respectively. The functions $FQ^{-}(f_i)$ and $FQ^{-}(f_i, f_j)$ are defined analogously for the negative instances. 


A key observation is that based on $FQ(f_i)$ and $FQ(f_i, f_j)$, we may compute $FQ(\overline{f_i})$, $FQ(f_i, \overline{f_j})$, $FQ(\overline{f_i}, f_j)$, and $FQ(\overline{f_i}, \overline{f_j})$. This is done as follows:

\begin{equation}
\label{equation:FQ1}
FQ^+(\overline{f_i}) = |\mathcal{D^+}| - FQ^+(f_{i})
\end{equation}

\begin{equation}
FQ^+(f_i, \overline{f_j}) = FQ^+(f_{i}) - FQ^+(f_{i}, f_{j})
\end{equation}

\begin{equation}
FQ^+(\overline{f_i}, f_j) = FQ^+(f_{j}) - FQ^+(f_{i}, f_{j})
\end{equation}

\begin{equation}
\label{equation:FQ4}
FQ^+(\overline{f_i}, \overline{f_j}) = |\mathcal{D^+}| - FQ^+(f_{i}) - FQ^+(f_{j}) + FQ^+(f_{i}, f_{j})
\end{equation}

The equations make use of the fact that the features are binary. For example, Eq.~\ref{equation:FQ1} states that if the total number of positive instances is $|\mathcal{D^+}|$ and we computed the frequency count $FQ^+(f_i)$, then the frequency count $FQ^+(\overline{f_i})$ is the number of instances in which $f_i$ does not appear, i.e., the difference between $|\mathcal{D^+}|$ and $FQ^+(f_i)$. Similar reasoning is applied to the other equations and computing the frequency count $FQ^-$ is analogous.

The following proposition summarises the runtime of computing $FQ^+(f_i, f_j)$.

\begin{proposition}(Computational Complexity of Phase One)
\label{prop:phaseone}
Let $m_+$ denote the maximum number of features in any single positive instance. Frequency counts $FQ^+(f_i, f_j)$ can be computed in $\mathcal{O}(|D^+| \cdot m^2_+)$ time with $\mathcal{O}(\mathcal{F}^2)$ memory.
\end{proposition}

An efficient way of computing the frequency counts is to represent the feature vector as a \emph{sparse} vector, and iterate through each instance in the dataset and increase a counter for each individual feature and each pair of features. This leads to the proposed complexity result. The additional memory is required to store the frequency counters, allowing to query a frequency count as a constant time operation. Note that the pairwise frequency count is symmetric, i.e., $FQ^+(f_i, f_j) = FQ^+(f_j, f_i)$, which requires only to consider $f_i$ and $f_j$ in the frequency count for $i < j$. This results in a smaller hidden constant in the big-O notation.

\subsubsection{Phase Two: Optimal tree computation (Algorithm \ref{algorithm:specialisedDT}, Lines \ref{specialisedtree:PhaseTwoStart}-\ref{specialisedtree:PhaseTwoEnd})}

Recall that a classification node is assigned the positive class if the number of positive instances exceeds the number of negative instances, otherwise the node class is negative. Let $CS(f_i, f_j)$ be the classification score for a classification node with all instances of $\mathcal{D}$ 
containing both features $f_i$ and $f_j$. The classification score is then computed as follows. 

\begin{equation}
\label{equation:CS}
CS(f_i, f_j) = \min\left\{FQ^+(f_i, f_j), FQ^-(f_i, f_j)\right\}
\end{equation}

Given a decision tree with depth two, a root node with feature $f_{root}$, a left and right children with features $f_{left}$ and $f_{right}$, we may compute the misclassification score in constant time assuming the frequency counts are available. Let $MS_{left}$ and $MS_{right}$ denote the misclassification scores of the left or right subtree. The computations are as follows.

\begin{equation}
\label{eq:specialisedMS1}
MS_{left}(f_{root}, f_{left}) = CS(\overline{f_{root}}, \overline{f_{left}}) + CS(\overline{f_{root}}, f_{left})
\end{equation}

\begin{equation}
\label{eq:specialisedMS2}
MS_{right}(f_{root}, f_{right}) = CS(f_{root}, \overline{f_{right}}) + CS(f_{root}, f_{right})
\end{equation}

The total misclassification score of the tree is the sum of misclassifications of its children. As the number of misclassification can be computed solely based on the frequency counts, we may conclude the computational complexity.

\begin{proposition}(Computational Complexity of Phase Two)
\label{prop:phasetwo}
Given the frequency counts $FQ^+$ and $FQ^-$, the optimal subtree tree can be computed in $\mathcal{O}(|F|^2)$ time with $\mathcal{O}(|F|)$ memory.
\end{proposition}

It follows from Property \ref{property:independence} that given a root node with feature $f_{root}$, the left and right subtrees can be optimised independently. Therefore, it is sufficient to compute for each feature its best left and right subtrees, and take the feature with the minimum sum of its child misclassifications. To compute the best left and right feature for each feature, the algorithm maintains information about the best left and right child for each feature found so far, leading to the memory requirement from Proposition \ref{prop:phasetwo}. The best features are initially arbitrarily chosen. Recall that from Property \ref{property:independence} it follows that the left and right subtree can be optimised independently:
$$
\min_{f_{left}, f_{right} \in \mathcal{F}} 
MS(f_{root},f_{left},f_{right}) = 
\min_{f_{left} \in \mathcal{F}} MS_{left}(f_{root},f_{left})
+ \min_{f_{right} \in \mathcal{F}} MS_{right}(f_{root},f_{right})
$$

Therefore, rather than considering triplets of features $(f_{root}, f_{left}, f_{right})$, it iterates through each pair of features $(f_{root}, f_{child})$, computes the misclassification values of the left subtree using Eq. \ref{eq:specialisedMS1}, updates the best left child for feature $f_{root}$, and performs the same procedure for the right child. After iterating through all pairs of features, the best left and right subtree is known for each feature, leading to the proposed complexity. The optimal decision tree can then be computed by finding the feature with minimum misclassification cost of its combined left and right misclassification.

After discussing each individual phase, we may conclude the overall complexity:

\begin{proposition}(Computational Complexity of Depth-2 Decision Trees)
\label{proposition:complexityDepth2}
Let $m$ be the upper limit on the number of features in any single positive and negative instance. The number of operations required to computing an optimal decision tree is $\mathcal{O}(|D| \cdot m^2 + |\mathcal{F}|^2)$ using $\mathcal{O}(\mathcal{F}^2)$ auxiliary memory. 
\end{proposition}

The result follows by combining Propositions \ref{prop:phaseone} and \ref{prop:phasetwo}. The obtained runtime is substantially lower at the expense of using additional memory compared to the dynamic programming formulation (Eq. \ref{DPformulation}) outlined in Proposition \ref{prop:generaldepth2}. Note that instances with binary features are naturally sparse. If the majority of instances contain more than half of the features, then as a preprocessing step all feature values may be inverted to achieve sparsity without loss of generality. The advantage of our approach is exemplified with lower sparsity ratios, i.e., cases where each vector contains a small number of features compared to the total number of features.

There are several additional points to note, which are not shown in Algorithm \ref{algorithm:specialisedDT} to keep the pseudo-code succinct. 

The above discussion assumed the feature node limit was set to three. The algorithm can be modified for the case of two feature nodes, keeping the same complexity, while in the case with only one feature node the pairwise computations are no longer necessary leading to $\mathcal{O}(|D| \cdot m + |F|)$ complexity, where $m$ is the upper limit on the number of features in any single positive and negative instance. As an optimisation technique, each time the algorithm is invoked, we extract the solutions using one, two, and three nodes and store all of these in cache (see Section \ref{section:caching}), regardless of the initial node count. The reasoning is that it is likely the other node counts will be considered in the future and the extra computation performed to capture all solutions is negligible. Furthermore, the algorithm is implemented to lexicographically minimise the misclassifications and the number of nodes.

To improve the performance in practice, the algorithm iterates through pairs of features $(f_i, f_j)$ such that $i < j$. After updating the current best left and right subtree feature using $f_i$ as the root and $f_j$ as the child, the same computation is done using $f_j$ as the root and $f_i$ as the child. Compared to the pseudo-code in Algorithm \ref{algorithm:specialisedDT}, this cuts the number of iterations by half, but each iteration does twice as much work, which overall results in a speed-up in practice. Moreover, rather than computing the best tree in a separate loop after computing the best left and right subtrees for each feature, this is done on the fly by keeping track of the best subtree encountered so far during the algorithm.

\emph{Specialised algorithm for decision trees of depth three}. We considered computing decision trees with depth three using a similar idea. Even though this results in a better big-O complexity for trees of depth three, albeit requiring $\mathcal{O}(\mathcal{F}^3)$ memory, our preliminary results did not indicate practical benefits. Including additional low-level optimisation might improve the results, but for the time being we leave this as an open question.

\subsubsection{Incremental Computation}

\label{section:incrementalFq}
The specialised method for computing decision trees of depth two is repeatedly called in the algorithm. For each call, the algorithm is given a different dataset that is a result of applying a split in one of the nodes in the tree. The key observation is that datasets which differ only in a small number of instances result in \emph{similar} frequency counts. The idea is to exploit this by only updating the necessary difference rather than recomputing the frequency counts from scratch. Such a strategy resembles techniques used in the frequent itemset community (\cite{diffsets}). We further elaborate on the idea used in our algorithm.

The key point is to view the previous dataset $\mathcal{D}_{old}$ and the new dataset $\mathcal{D}_{new}$ in terms of their intersection and differences. 

\begin{observation}
Given two datasets $\mathcal{D}_{new}$ and $\mathcal{D}_{old}$, let their difference be denoted as $\mathcal{D}_{in} = \mathcal{D}_{new} \setminus \mathcal{D}_{old}$ and $\mathcal{D}_{out} = \mathcal{D}_{old} \setminus \mathcal{D}_{new}$ and their intersection as $\mathcal{D}_{same} = \mathcal{D}_{new} \cap \mathcal{D}_{old}$. We may express the datasets as $\mathcal{D}_{new} = \mathcal{D}_{in} \cup \mathcal{D}_{same}$ and $\mathcal{D}_{old} = \mathcal{D}_{out} \cup \mathcal{D}_{same}$
\end{observation}

We first note that set operations can be done efficiently for datasets.

\begin{proposition}(Computational Complexity of Set Operations on Datasets)
\label{proposition:setoperations}
Given a dataset $\mathcal{D}$ and two of its subsets $\mathcal{D}_{new} \subseteq \mathcal{D}$ and $\mathcal{D}_{old} \subseteq \mathcal{D}$, the sets $\mathcal{D}_{in} = \mathcal{D}_{new} - \mathcal{D}_{old}$ and $\mathcal{D}_{out} = \mathcal{D}_{old} - \mathcal{D}_{new}$ can be computed in $\mathcal{O}(|\mathcal{D}_{new}| + |\mathcal{D}_{old}|)$ time using $\mathcal{O}(|\mathcal{D}|)$ memory.
\end{proposition}

The above can be realised by associating each instance of the original dataset $\mathcal{D}$ with a unique ID and storing positive and negative instances in their corresponding positive and negative arrays sorted by ID. Once these conditions are met, a linear pass through the datasets may determine the differences, and accordingly the frequency counts may be updated \emph{incrementally}.

\begin{proposition}(Computational Complexity of Incremental Frequency Computation)
\label{prop:incComplexity}
Let $m$ denote the maximum number of features in any considered instance. Given the frequency counts $FQ_{old}$ of a previous dataset $\mathcal{D}_{old}$, a new dataset $\mathcal{D}_{new}$, and their differences $\mathcal{D}_{in}$ and $\mathcal{D}_{out}$, the frequency counts $FQ_{new}$ of the new dataset $\mathcal{D}_{new}$ can be computed in $\mathcal{O}((|\mathcal{D}_{in}| + |\mathcal{D}_{out}|) \cdot m^2)$ time.
\end{proposition}

To show the complexity, note the difference between $FQ_{old}$ and $FQ_{new}$. 

\begin{observation}
\label{observation:K}
Let $\mathcal{K}(FQ)$ denote the set of instances used to compute the frequency counts $FQ$. It follows that $\mathcal{K}(FQ_{old}) = \mathcal{D}_{out} \cup \mathcal{D}_{same}$  and $\mathcal{K}(FQ_{new}) = \mathcal{D}_{in} \cup \mathcal{D}_{same}$.
\end{observation}

Consider taking $FQ_{old}$ and applying a series of operations to reach the new frequency counts $FQ_{new}$. The complexity result of Proposition \ref{prop:incComplexity} follows from the previous observations and the following:

\begin{observation}
The frequency counts $FQ_{old}$ already capture the counts for instances $\mathcal{D}_{same}$
\end{observation}

\begin{observation}
The frequency counts $FQ_{old}$ need to be incremented using instances $\mathcal{D}_{in}$
\end{observation}

\begin{observation}
The frequency counts $FQ_{old}$ need to be decremented using instances $\mathcal{D}_{out}$
\end{observation}

Using the incremental update procedure is sensible only if the number of updates required is small compared to recomputing from scratch. In our algorithm, in each call to compute a decision tree of depth two, the algorithm incurs an overhead (Proposition \ref{proposition:setoperations}) to compute the differences between the old and new dataset. It proceeds with the incremental computation if $|\mathcal{D}_{in} \cup \mathcal{D}_{out}| < |\mathcal{D}_{new}|$, and otherwise computes from scratch. 

Our algorithm keeps two sets of frequency counters, which are tied to two different datasets $\mathcal{D}_{old}^1$ and $\mathcal{D}_{old}^2$. When our method is called, the algorithm uses as a starting point the previous frequency counter that requires the least number of operations to incrementally construct the new new frequency counter for the new dataset $\mathcal{D}_{new}$. Upon completing the frequency counter computation, the new counter will replace the chosen old one.

The overhead of computing the number of operations required is negligible compared to the overall complexity of computing the optimal tree of depth two (Proposition \ref{proposition:complexityDepth2}), but the benefits can be significant if the difference is small. Assuming that two neighbouring features are similar, two successive features considered for splitting are likely to lead to require only a small number of modifications.

The previous paragraph motivates the choice of only storing \emph{two} frequency counters. When computing a tree of depth three, the data is split amongst the left and right subtree. Whereas the data passed to the left and right subtree may be very different from one another, the data passed to the left subtree during the \emph{next} split may not be substantially different from the data used in current split in the left subtree (similarly for the right subtree). Recall that the specialised depth two algorithm would be called on the child subtree sequentially. In order to preserve the frequency counters of both children in between two successive splits, the heuristic choice was made to store two sets of frequency counters. As shown in the experimental section, the incremental strategy provides notable runtime reductions.

\subsection{Similarity-Based Lower Bounding}
\label{section:similarityBounding}

We present a novel lower bounding technique that does not rely on the algorithm having previously searched a specific path, as opposed to the cache-based lower bound introduced in the later section. Given a dataset $\mathcal{D}_{new}$ for a node, the method aims to derive a lower bound by taking into account a previously computed optimal decision tree using the dataset $\mathcal{D}_{old}$. It infers the bound by considering the difference in the number of instances between the previous dataset $\mathcal{D}_{old}$ and the current dataset $\mathcal{D}_{new}$. The bound is used to prune portions of the search space that are guaranteed to not contain a better solution than the best decision tree encountered so far in the search. We note that our approach is related to the lower bound for decision lists (\cite{angelino2017learning}) and \emph{diffset} computation (\cite{diffsets}). We present the ideas as an application to decision trees using elementary algebra.

Assume that for both datasets, the depth and the number of allowed feature nodes requirements are identical. As in the previous section, we define the sets $\mathcal{D}_{in} = \mathcal{D}_{new} \setminus \mathcal{D}_{old}$, $\mathcal{D}_{out} = \mathcal{D}_{old} \setminus \mathcal{D}_{new}$, and $\mathcal{D}_{same} = \mathcal{D}_{new} \cap \mathcal{D}_{old}$.

Given the limit on the depth $d$ and number of features nodes $n$, a dataset $\mathcal{D}_{new}$, and a dataset $\mathcal{D}_{old}$ with $T(\mathcal{D}_{old}, d, n)$ as the misclassification score of the optimal decision tree of $\mathcal{D}_{old}$ (recall Eq. \ref{DPformulation}), we define the similarity-based lower bound,

\begin{equation}
\label{eq:similarLB}
LB(\mathcal{D}_{new}, \mathcal{D}_{old}, d, n) = T(\mathcal{D}_{old}, d, n) - |\mathcal{D}_{out}|,
\end{equation}

\noindent
which is a lower bound for the number of misclassifications of the optimal decision tree for the dataset $\mathcal{D}_{new}$ of a tree of depth $d$ with $n$ feature nodes. Formally: 

\begin{proposition}
\label{proposition:LB}
$LB(\mathcal{D}_{new}, \mathcal{D}_{old}, d, n) \leq T(\mathcal{D}_{new}, d, n)$.
\end{proposition}

As a result, subtrees with a lower bound greater than its upper bound are immediately pruned, effectively speeding up the search. To show that Proposition \ref{proposition:LB} is indeed a lower bound, let $T(\mathcal{D}) = T(\mathcal{D}, d, n)$, note that removing $\mathcal{D}_{out}$ from $\mathcal{D}_{old}$ may reduce the misclassification cost by at most $|\mathcal{D}_{out}|$:

\begin{equation}
T(\mathcal{D}_{old}) - T(\mathcal{D}_{old} \setminus \mathcal{D}_{out}) = 
T(\mathcal{D}_{old}) - T(\mathcal{D}_{same})
\leq |\mathcal{D}_{out}|.
\end{equation}

\begin{equation}
\label{eq:limiteddecrease}
T(\mathcal{D}_{old}) - |\mathcal{D}_{out}|
\leq T(\mathcal{D}_{same}).
\end{equation}

Adding instances to $\mathcal{D}_{same}$ cannot decrease the misclassification score $T(\mathcal{D}_{same})$:

\begin{equation}
\label{eq:nondecreasing}
T(\mathcal{D}_{new}) = T(\mathcal{D}_{same} \cup \mathcal{D}_{in}) \geq T(\mathcal{D}_{same}) 
\end{equation}

Combining Eq. \ref{eq:limiteddecrease} and \ref{eq:nondecreasing} we arrive at:

\begin{equation}
T(\mathcal{D}_{old}) - |\mathcal{D}_{out}| \leq T(\mathcal{D}_{new})
\end{equation}

\begin{equation}
LB(\mathcal{D}_{new}, \mathcal{D}_{old}, d, n) \leq T(\mathcal{D}_{new})
\end{equation}

\noindent
which shows the derivation of Proposition \ref{proposition:LB}.

As implied in the previous text, a set of previous datasets and their optimal values need to be kept available for comparison once a new dataset is considered. This give rise to a trade-off: keeping more datasets increases the probability of deriving a greater lower bound at the expense of computational time for each lower bound computation. 

Our algorithm stores two datasets for each depth value. When computing the bound for a new dataset with depth $d$, the two datasets stored at depth $d$ are used to compute the similarity-based lower bound, and the stronger bound of two is taken. Once a subtree has been exhaustively searched with depth $d$, its corresponding dataset replaces the most similar dataset stored at depth $d$. Similarity between datasets is formally computed as $|\mathcal{D}_{out}|+|\mathcal{D}_{in}|$. The intuition is to ensure that given two successive splits at depth $d$, the resulting left and right subtree datasets of the first split will be used to compute the bound of the dataset that come in the next split. This is similar reasoning as in the case of the incremental computation in the specialised algorithm (see end of Section \ref{section:incrementalFq}). 

If the datasets $\mathcal{D}_{old}$ and $\mathcal{D}_{new}$ are equal, then any result for dataset $\mathcal{D}_{old}$ can be directly used for $\mathcal{D}_{new}$. As discussed in the next section, optimal solutions and lower bounds for subtrees are stored in the cache. When computing a similarity-based lower bound for a new dataset at depth $d$, if it is detected that it is equal to one of the two stored datasets at depth $d$, then the optimal solution and lower bounds of the stored datasets are fully transferred to the new dataset. Note that this situation may only occur when using a branch-based caching (see next section).

As shown in the experimental results (Section \ref{section:partZwei}), the use of the similarity-based lower bound reduces the runtime for all datasets, with only a few exceptions.

\subsection{Caching of Optimal Subtrees (Memoisation)}
\label{section:caching}	
	
As is common in dynamic programming algorithms, a caching or memoisation table is maintained to avoid recomputing subproblems. In our case, information about computed optimal subtrees is stored. This is used to retrieve a subtree that has already been computed, provide lower bounds, and reconstruct the optimal decision tree at the end of the algorithm. Caching has been used in previous works (\cite{dl8,nijssen2010optimal,dl85,verhaeghe2019learning,hu2019optimal,sparseICML}). 

We adapt and extend the caching techniques from the literature for our algorithm, i.e., our cache takes into account constraints both on depth and the number of nodes. We discuss two caching techniques, namely \emph{branch} and \emph{dataset} caching, that have been introduced in DL8 (\cite{dl8,nijssen2010optimal}) and (variants) have appeared in later works. Whereas the different techniques were viewed as a trade-off between computational time and efficiency, we show that in our realisation both techniques take only a small fraction of the total runtime, allowing us to use dataset caching without incurring notably drawbacks.

Formally, we define the cache as a mapping of a subtree (represented as a branch or dataset, see below) to a set of \emph{cache entries}. Each entry contains information on the lower bound and/or the optimal root node of the subtree under constraints on the depth and number of feature nodes, which includes the root feature, the number of feature nodes in its left and right children, and the misclassification score. Initially, the cache is empty and is populated throughout the algorithm. As we employ a specialised algorithm for depth two trees, we do not cache the lowest decision tree layer. This leads to fewer subtree entries in the cache, saving space and increasing efficiency. We note that given our techniques described below, the overhead of look-up a subtree in our cache is kept low.

A key concept is the hash function of a subtree. We discuss the branch and dataset representations, corresponding hash functions, and our realisation of these ideas.

\subsubsection{Subtree Hashing Based on Branches}

The key observation is that given a path from the root to any given node,
each permutation of the feature nodes on the path results in the same
dataset for the node furthest from the root, e.g.,
$\mathcal{D}(f_i)(\overline{f_j}) = \mathcal{D}(\overline{f_j})(f_i)$. This allows representing a path as a set of features, e.g., $\{f_i,
\overline{f_j}\}$. The path of a subtree is called a \emph{branch}. We reiterate that each subtree may be associated with exactly one branch, while a single branch of length $k$ may be linked to $k!$ subtrees. This is valuable since it implies that the computation of a single subtree may be shared amongst each $k!$ subtree associated with the same branch. We note that the \emph{branch} representation has been introduced in DL8 using the term \emph{itemset}.

Our branch-based cache consists of an array of hash tables: each branch of length $k$ is stored in the \emph{k-th} hash table. We stored the branch as a sorted array, where features are converted into integers based on their index and polarity\footnote{Such a conversion is borrowed from the SAT solving literature.}, i.e., given a feature $f$ with index $i$, we assign the value $2i+1$ to the positive feature $f$ and $2i$ to the negative feature $\overline{f}$. We use a conventional hash function\footnote{E.g., see the function template \emph{hash\_combine} in the C++ Boost library.} on integer arrays within the hash tables, i.e., given an array $A$ of length $n$, its hash is computed using Algorithm \ref{alg:hash}.

\input{algo_hashing}

The advantage of a branch representation is its compactness, i.e., each subtree is represented only using a few features, allowing efficient hash function computation. The downside is that different branches may lead to the same dataset (subproblem), but this will not be detected when using caching based on branches, leading to unnecessary recomputation.

\subsubsection{Subtree Hashing Based on Datasets}

An alternative to the branch representation is to use more general representations. This is desirable since once a lower bound or optimal solution has been computed for a subtree, the results may be shared amongst any future subtree that optimises exactly the same subproblem rather than only sharing with subtrees with equivalent branches, alleviating the drawback of the branch representation. The downside is that more general representations may be computationally intensive.

DL8 (\cite{dl8,nijssen2010optimal}) proposed to compute the \emph{closure} of a branch (itemset): given a branch, its closure is the set of features that appear in all instances of the dataset corresponding to the branch. The closure is then used as the subproblem representation. Note that the branch is a subset of its closure. Such an approach correctly identifies all equivalent subproblems, addressing the issue of the branch representation. The drawback is that computing the closure requires additional computation, providing an important trade-off that must be considered. A related idea has been recently proposed by \cite{sparseICML}, where the subproblem is represented using a bitset, i.e., the \emph{i-th} bit is set is the \emph{i-th} instance is present.

In our work, we introduce an alternative representation that uses the dataset as the subproblem representation and discuss several techniques that keep the computational time of caching low.

At the start of the algorithm, each instance is assigned a unique identifier in the form of an integer. A dataset contains an array for each class and instances within each array are kept sorted with respect to their identifier. Given such data structures, determining whether two datasets $\mathcal{D}_1$ and $\mathcal{D}_2$ are identical may be done in linear time with respect to the number of present instances, i.e., $\mathcal{O}(|\mathcal{D}_1|)$.

Our cache consists of an array of hash tables, where datasets with $m$ instances are stored in the \emph{m-th} hash table. We use instance identifiers and Algorithm \ref{alg:hash} to compute the hash value of a dataset, where the dataset is treated as an array with instance identifiers as integers. The hash is computed only once and stored in the dataset for further use. 

In our experiments, the effectiveness of dataset caching showed better performance than branch caching and the additional memory requirements were not an issue.

For both caching ideas we introduced a further optimisation to speed-up in practice. For each array of array of hash tables, the last two calls are stored in a temporary buffer. Before searching in the hash table, the cache first looks up the temporary buffer in case the item sought for is already present. This was done since the same cache call may be performed many times during the algorithm, e.g., after splitting the dataset, the algorithm select the node allocation to the left and right child, but each cache call will lead to the same cache entry.

\subsubsection{Storing Subtrees and Lower Bounds in the Cache}
\label{section:LBstore}


Information is stored in cache once a subtree has been exhaustively explored. We consider two scenarios:

\emph{Scenario \#1: a decision tree has been found within the upper bound}. In this situation, the computed subtree is optimal and the corresponding entry is stored/updated using the root node assignment as the optimal assignment, the lower bound is set to the misclassification score, noting the depth and feature node limit that was given when computing the subtree. In the event that the algorithm determines that the minimum misclassification score may be achieved using fewer nodes than imposed by the node limit, we may use the following proposition to create additional cache entries:

\begin{proposition}
Let $T(\mathcal{D}, d, n)$ be the misclassification score of the optimal
decision tree for the dataset $\mathcal{D}$ with depth limit $d$ and node
limit $n$. If there exists an $n' < n$ such that $T(\mathcal{D}, d,n') =
T(\mathcal{D}, d,n)$, then $T(\mathcal{D}, d, i) = T(\mathcal{D}, d, n)$ for $i \in [n', n]$.   
\end{proposition} 

Similar reasoning is used to populate entries in case a smaller depth is used than allowed. We note that during the algorithm, a given branch or dataset may be only exhaustively explored once, depending on the subtree representation used in the cache. The next time a branch or dataset is encountered, its corresponding solution is retrieved from the cache (Section \ref{section:LBcache}). 

\emph{Scenario two: no decision tree has been found within the upper bound}. It follows that the lower bound on the number of misclassifications for the subtree is at least one greater than the upper bound. This is the lower bounding reasoning introduced in DL8.5 (\cite{dl85}). 

In this work, we note a slightly stronger lower bound. Let $LB(\mathcal{D}, d, n)$ be the lower bound for the number of misclassifications of an optimal decision tree for the dataset $\mathcal{D}$ with \emph{n} nodes and depth \emph{d}, i.e., $T(\mathcal{D}, d, n) \geq LB(\mathcal{D}, d, n)$. We introduce the following refined lower bound $RLB$:

\begin{equation}
\label{eq:localBound}
LB_{local}(\mathcal{D}, f, d, n_{left}, n_{right}) = LB(\mathcal{D}(\overline{f}, d-1, n_{left})) + LB(\mathcal{D}(f, d-1, n_{right}))
\end{equation}

\begin{equation}
\label{eq:slb}
RLB(\mathcal{D}, d, n) = \min\{LB_{local}(\mathcal{D}, f, d, n_{left}, n_{right})\ |\ f \in \mathcal{F}\ \land\ n_{left} + n_{right} = n-1\}
\end{equation}

The bound $RLB$ considers all possible assignments of features and numbers of feature nodes to the root and its children, and selects the minimum sum of the lower bounds of its children. It follows that no decision tree may have a misclassification score lower than $RLB$. We combine $RLB$ with the upper bound to obtain a lower bound for the case where no decision tree with less than the specified upper bound $UB$ could be found:

\begin{equation}
\label{eq:generalisedDL85bound}
T(\mathcal{D}, d, n) \geq max\{RLB(\mathcal{D}, d, n),\ UB+1\}. 
\end{equation}

The proposed bound generalises the bound from DL8.5 (\cite{dl85}), which only considers the second expression on the right-hand side of Eq. \ref{eq:generalisedDL85bound} to derive a lower bound when no tree could be found within the given upper bound. 

Once the lower bound has been computed, it is recorded in the cache for the subtree along with the constraints on the depth and number of feature nodes, and the optimal assignment is set to null. 

\subsubsection{Retrieving Subtrees and Lower Bounds from the Cache}
\label{section:LBcache}

When considering a new subtree, the algorithm searches for set of entries corresponding to the subtree using hash tables, as discussed at the beginning of the section.

A lower bound for the current tree may be inferred from the bounds of the larger tree, formally summarised in the following proposition. 

\begin{proposition}
\label{prop:bigtreebound}
Given the dataset $\mathcal{D}$ and depth bound $d$ and the maximum number
of feature nodes $n$, 
a bound for a larger tree is a bound for the current tree, i.e., $\forall n'
\geq n, d' \geq d: LB(\mathcal{D}, d', n') \leq LB(\mathcal{D}, d, n)$.
\end{proposition}

When retrieving a lower bound for trees that have no cache entries, Proposition \ref{prop:bigtreebound} allows inferring a lower bound from larger trees that may be stored in the cache. Note that the lower bounds are nonincreasing with size, i.e.,

\begin{equation}
LB(\mathcal{D}, d', n') \leq LB(\mathcal{D}, d, n) \qquad n' \geq n \land d' \geq d
\end{equation}

The tightest bound is returned when retrieving the lower bound. If there are no applicable entries in the cache, the trivial lower bound of zero is returned. For example, given a dataset $\mathcal{D}(f_1, f_2)$ with the node limit set to five and depth three, if there is no subtree for the given size in the cache but there is an entry when the node limit was set to six and seven with the same depth, then the lower bound using six nodes is the tightest valid lower bound available for the tree with five nodes.

\subsubsection{Incremental Solving}
\label{section:incrementalsolving}

We label the process of querying the algorithm to compute progressively smaller or larger decision trees as \emph{incremental solving}. For example, once the algorithm has computed an optimal decision tree for a given depth and number of nodes, the user may be interested in a tree with more or fewer nodes. Such situations also occur as part of hyper-parameter tuning. Our cache naturally supports these types of queries. Computations used for a tree with a given depth and node count may be reused when the algorithm is run with a different set of depth and node count values.

\subsubsection{Recovering the Optimal Decision Tree}

Recall that for each subtree optimally computed, only the root node is stored in the cache. When necessary, the complete subtree may be reconstructed as a series of queries to the cache, where each time a single node is retrieved, as introduced in DL8 (\cite{dl8,nijssen2010optimal}). In our algorithm, there is an exception to the mentioned tree reconstruction procedure. After solving a tree of depth two, the root node is stored, but not its children. During the algorithm these are not necessary, but the children are needed when reconstructing the optimal decision tree found at the end. In this case, the required child nodes are recomputed using Algorithm \ref{algorithm:specialisedDT}. This avoids storing an exponential number of entries (recall that the number of paths increases exponentially with the depth) which do not serve a purpose other than the final reconstruction. The resulting computational overhead of recomputing the solutions at the end is negligible compared to the overall execution time.

\subsection{Node Selection Strategy}
\label{section:nodeselection}

Given a feature for a node and the size allocation for its children, the
algorithm decides on which child node to recurse on first. 
In DL8.5, the algorithm always visits the left subtree before the right subtree. 

Our search strategy is a variant of such post-order traversal, labelled \emph{dynamic ordering}, which dynamically decides which child node to visit
first. The idea is to prioritise the child node that has (heuristically) the higher number of misclassifications, which in turn leads to a higher chance to prune to the other sibling. The potential is roughly approximated by the number of misclassifications of its corresponding classification node. In our experiments such a strategy shows consistent improvement over a static post-order traversal used in DL8.5 (\cite{dl85}). We considered a more complex approach that selected the subtree with the larger gap between the upper and lower bound, however this did not leave to improvements over the dynamic strategy ordering.

Ordering search nodes according to a heuristic is common in combinatorial optimisation, e.g., variable selection in integer/constraint programming, and in the data mining literature, e.g., \cite{zaki2000scalable,zimmermann2009cluster}. The above idea represents such an idea applied to optimal decision trees. 

\subsection{Feature Selection}
\label{section:assignments}

For a given node, each possible tree configuration (a feature and the size of its children) is considered one at a time, unless the node is pruned or the optimal solution is retrieved from the cache (see Subsection \ref{section:caching}). The order in which tree configurations are explored may have an impact on performance, as evidenced in search algorithms in general. 

We considered three feature selection strategies from the literature, which order the features according to the following: 1) Gini coefficient of the features, 2) in order of feature appearance in the dataset, and 3) randomly order features. As discussed in (Section \ref{section:experiments:partthree}), in our experiments the inorder variant was selected as the default option.

\subsection{Extensions}
\label{section:extensions}

The core algorithm has been presented in the previous sections. We now discuss several extensions that use the presented core algorithm as a basis.

\subsubsection{Multi-Classification}

To extend the algorithm for multi-classification, the key step is to generalise Algorithm \ref{algorithm:specialisedDT} to compute frequency counters for each class. Equations analogous to Equations \ref{equation:FQ1}---\ref{eq:specialisedMS2} are devised to compute the misclassification scores. Since classes partition the data, the complexity results remain valid for multi-classification.

\subsubsection{Regression}

As in multi-classification, the main step in extending our method to work with regression is to adapt the specialised algorithm for computing depth two trees (Algorithm \ref{algorithm:specialisedDT}). Consider regression trees where leaf nodes are assigned fixed values that are computed as the average value of their corresponding training instances. The specialised algorithm for depth two trees, in addition to the frequency counters, maintains a similar structure where the total sum of target values of each pair of features is stored, and analogous equations to Equations \ref{equation:FQ1}---\ref{eq:specialisedMS2} are used. Note that our similarity-based lower bounds, in their current form, would not be applicable to regression.

\subsubsection{Sparse Objective}

Apart from minimising the misclassification score, the \emph{sparse objective} is a popular metric for decision trees, which considers a weighted linear combination of the misclassification score and the number of feature nodes. This objective was used in the original CART paper (\cite{breiman1984classification}) and discussed in some of the other optimal decision tree works (\cite{bertsimas2017optimal,hu2019optimal,sparseICML}). Formally, the sparse objective is specified as follows: 

\begin{equation}
\label{eq:sparseobjective}
misclassifications + \alpha \times nodes,
\end{equation}

\noindent
balancing the size of the decision tree against the misclassifications using the \emph{sparse coefficient} $\alpha \in \mathbb{N}_{0}$. The intuition is that adding a node to a decision tree should only be considered if it leads to a reduction in the misclassifications by at least $\alpha$. The depth may also be penalised in a similar manner.

To support the sparse objective, we perform a sequence of queries to our algorithm, each time modifying the number of nodes and imposing an upper bound according to the new number of nodes considered, illustrated in Algorithm \ref{alg:SparseObjective}. Given our caching mechanism, computations in between calls are reused, even if the sparse weight $\alpha$ is changed (see below). 

\input{algo_sparse_objective}

Note that, as an emerging functionality, running Algorithm \ref{alg:SparseObjective} with $\alpha=0$ considers all possible trees given the upper values on the depth and number of nodes and stores the computation in the cache. As a result, the optimal tree given \emph{any} $\alpha$ may be computed immediately, as all the necessary subtrees needed for the computation will be already stored in the cache. This may be considered when tuning a tree for the best sparse coefficient $\alpha$. In this case, it is important that the depth is small enough that the algorithm may terminate within reasonable time, as the runtime exponentially increases with respect to the depth. Otherwise, for larger depths, it may be beneficial to consider particular $\alpha$ values rather than tuning for all values, since setting the coefficient to a positive value $\alpha > 0$ contributes towards reducing the search space, which may assist the search in the case of deeper trees. 



\subsubsection{Minimising Number of Nodes}

We may consider a lexicographical objective, where the aim is to achieve the minimum misclassification score using the least amount of nodes. Note that our presented algorithm is focussed on minimising misclassifications within the given constraints on the depth and number of nodes, but does not necessarily return the smallest tree. To take into account the lexicographical objective, we may first compute the optimal tree, and then query the algorithm to compute smaller trees using the misclassification score as an upper bound. Recall that computations from one call will be reused in other calls through the cache. This is summarised in Algorithm \ref{alg:minNodes}.

\input{algo_minimising_num_nodes}


\subsubsection{Anytime Behaviour}

An \emph{anytime} algorithm has the property that it may provide a solution at any time during its execution. Our algorithm discussed in the previous sections has, in some sense, support for anytime behaviour since a global solution is only registered at the root node. However, the algorithm may be augmented to save a better solution as soon as it is found, rather than delay until the root node recursion. The key observation is that \emph{any} solution, even a classification node, is \emph{a} solution to the decision tree problem. When processing one child subtree, the other subtree may be assumed to be a classification node for the purposes of the anytime solution. A separate data structure may be maintained to keep track of the current tree, incrementally update the misclassification score, and update the global solution should a better solution present itself during search. This incurs an additional overhead, but it is negligible compared to the other operations, e.g., computing a tree of depth two. We note that while anytime behaviour may be supported, our algorithm is not designed to exhibit strong anytime behaviour, but rather minimise the time to exhaustively explore the search space of all decision trees. Another technique to improve the anytime behaviour is to consider an \emph{iterative deepening} approach, where the optimal trees of depth $k$ is computed before proceeding with trees of depth $k+1$ until the desired maximum depth is reached, possibly using Gini feature selection.

\subsubsection{Optimising Nonlinear Metrics}

Our algorithm may be extended to support metrics which have a nonlinear relationship between the misclassifications of each class, e.g., F1-score. The key idea is to generalise the method to a bi-objective setting, where the objectives represent the number of misclassified instances for each individual class. This allows computing the Pareto front and subsequently computing the tree that minimises the given nonlinear metric. While this is desirable, the task of minimises such metrics is significantly more computationally expensive compared to linear metrics such as the misclassification score. We refer the interested reader to a separate article (\cite{biObjTrees}), where we explored this idea in more detail.





\section{Computational Study}
\label{section:experiments}

The goal of this section is to evaluate different variants of our algorithm and compare with the state-of-the-art. With this in mind, we designed three major themes to investigate, each addressing a unique set of questions: 1) variations and scalability of our approach, 2) effectiveness compared to the state-of-the-art optimal classification tree algorithms, and 3) out-of-sample accuracy as compared to heuristically-obtained decision trees and random forests.

\subsection{Datasets and Computational Environment}

We use publicly available datasets used in previous works (\cite{bertsimas2017optimal,verwer2019learning,narodytska2018learning,dl85,hu2019optimal}), most of which are available from the UCI and CP4IM repositories. The datasets include 85 classification problems with a mixture of binary, categorical, and continuous features. Datasets with missing values were excluded from experimentation. Some benchmarks appeared in previous works under different binarisation techniques or simplifications, e.g., multi-classification turned into binary-classification using a `one-versus-all' scheme or a subset of the features were removed. We include such variants as separate datasets. 

Datasets with categorical and/or continuous features are converted into binary datasets as a  preprocessing step using a supervised discretisation algorithm based on the \emph{minimum description length principle} (MDLP) by \cite{DBLP:conf/ijcai/FayyadI93}, effectively converting each feature into a categorical feature based on the statistical significance of the feature values for the class, and then using a one-hot encoding to binarise the features. This was done for 22 datasets, while the remaining 63 datasets were already binary. The implementation of MDLP from the R programming package was used (\cite{kim2015package}). While the chosen discretisation strategy was sufficient for our evaluation purpose, we acknowledge that there may be better ways of discretising features. We leave the analysis of discretisation strategies for optimal decision trees for future work.

We partition the datasets into four groups based on the source of the dataset. For this reason the names displayed do not follow an alphabetical order.

Our code, binarised datasets, and the binarisation script are available online: \url{https://bitbucket.org/EmirD/murtree}.

Experiments were run on an Intel i-7-8550U CPU with 32 GB of RAM running one algorithm at a time using one processor. The timeout was set to ten minutes except for the hyper-parameter tuning where no limit was enforced. In the following, we dedicate a separate subsection to each of the three major experimental topics.

\subsection{Variations of Our Algorithm and Scalability}

The aim of this subsection is to investigate variations of our approach, namely:

\begin{enumerate}
\item Compare branching and dataset caching.
\item Assess the efficiency of incremental frequency and similarity lower bound computation.
\item Analyse the impact of the feature and node selection strategies.
\item Demonstrate the limits and scalability of our approach.
\end{enumerate}

The default setting of our algorithm uses all techniques presented in the paper, i.e., incremental frequency computation, incremental solving, similarity lower bounding, in-order feature selection, dynamic node selection, and the dataset-based cache.

\subsubsection{Part One: Cache Variants}
\label{section:partOne}

We run our algorithm varying the cache strategy (branch- and dataset-based caching: Section \ref{section:caching}). For each combination, we fix the depth of the tree to four and task the algorithms to compute fifteen optimal decision trees, one tree for each value of $n \in [1, 15]$. Such a computation task is common in hyper-parameter tuning or when optimising the sparse objective. The algorithms take advantage of incremental solving, i.e., subproblems stored in the cache from previous trees are reused. 

The results are given in Table \ref{table:hashComparison}, where the runtime and number of cache entries for each setting are shown. Benchmarks where the difference was insignificant are excluded.

\input{table_hash}

Based on the results, we conclude that the dataset cache leads to the best performance in terms of runtime. We note that the difference in efficiency is also reflected in the number of cache entries. Dataset caching requires fewer cache entries, indicating that the equivalence between difference subproblems could be effectively exploited. The exception are two datasets, where the number of cache entries is similar and branch caching has a slight advantage. The reason for only a slight difference is that even though dataset caching requires more computational time and memory compared to branch caching, the overall difference is not significant in our implementation compared to the other algorithmic components. We note that we experimented with a cache based on the closure of a branch, but this incurred a notable overhead for most instances compared to dataset- and branch-caching.

\subsubsection{Part Two: Incremental Frequency Computation and the Similarty-Based Lower Bound}
\label{section:partZwei}

We run our algorithm tuning on and off the incremental frequency computation (Section \ref{section:incrementalFq}) and the similarity-based lower bound (Section \ref{section:similarityBounding}). This gives rise to a total of four settings. As before, the depth of the tree is fixed to four and the algorithms compute fifteen optimal trees by varying the number of nodes from one to fifteen. 

The results are given in Table \ref{table:partZwei}, where the runtime is shown for each setting. Benchmarks where the difference was insignificant are excluded.

We draw two main conclusions. First, incremental computation is always beneficial. The splits of two neighbouring features often only differ in a small number of instances, and consequently performing minor changes to the previously computed frequency counters saves time compared to recomputing from scratch. Second, the lower bound contribute towards equal or lower runtimes in a most of the benchmarks. Overall, the experiments demonstrate that it is typically beneficial to include both incremental computation and lower bounding.

\input{table_our_techniques}

\subsubsection{Part Three: Feature and Node Selection Strategies}
\label{section:experiments:partthree}

We run our algorithm varying the feature selection strategy (Section \ref{section:assignments}: in order, random, and sorted according to Gini coefficients) and the node selection strategy (Section \ref{section:nodeselection}: post-order and dynamic). As before, the depth of the tree is fixed to four and the algorithms compute fifteen optimal trees by varying the number of nodes from one to fifteen. 

The results are given in Table \ref{table:partthree}, where the runtime is shown for each setting. Benchmarks where the difference was insignificant are excluded.

We draw several main conclusions. First, random feature selection is never beneficial. This is partially due to its anti-synergy with our incremental frequency computation and similarity lower bounding. Second, dynamic node selection is consistently better than a fixed post-order selection, albeit the performance gains are relatively small compared to the other algorithmic components. Third, both Gini and in-order feature selection are competitive, although in-order performs better on slightly more datasets while being much simpler.

\input{table_feature_node}

\subsubsection{Part Four: Scalability}

We investigate the sensitivity of our MurTree algorithm with respect to the number of instances and maximum depth. In Table \ref{table:scalability}, results are shown for a subset of representation datasets when our algorithm is run to compute trees of $depth \in [4, 5]$ on datasets where instances are duplicated $k \in [1, 2, 3, 4]$ times. Similarly as before, each run computes fifteen and thirty-one trees with varying size, depending on the depth. Depth three trees are computed in negligible time and are excluded from further consideration.

The results indicate a linear dependency with the number of instances for the majority of the datasets. As most of the computational time is spent in repeatedly solving optimal subtrees of depth two (Section \ref{section:specialisedAlgorithm}), the finding is consistent with the theoretical complexity (Proposition \ref{proposition:complexityDepth2}). This is a notable improvement over generic optimisation approaches, such as integer programming or SAT. The latter may exhibit an exponential runtime dependency on the number of instances as new binary variables are introduced for each instance, and typically do not consider datasets with more than a thousand instances. 

In contrast to the number of instances, the depth has a large impact on the running time. The number of possible decision trees grows exponentially as the depth increases, which is reflected in the sharp increase of both the time and number of cache entries. For example, our approach computes depth-three trees within seconds, but the runtimes go up notably for depth four and five, e.g., the differences between depths is in the order of magnitude. Our proposed techniques manage to delay the exponential blow up, but do not remove it.

The main conclusion of the above discussion is that the bottleneck of the approach is not necessarily in the number of instances, but rather in the depth of the tree. Note that the experiments are merely indicative. In practice, however, introducing more instances might implicitly increase or decrease the number of binary features in the discretisation and have an effect on shaping the structure of the dataset, both of which may impact positively or negatively the running time. 

Apart from the depth, another important factor is the number of binary features, which additionally dictates the number of possible decision trees necessary to explore to find the optimal tree. As the ability of our techniques to prune and reduce computational time depends on the structure of the dataset, it is difficult to artificially increase the number of features and show the dependency. For example, duplicating features would not lead to conclusive statements on the impact of the number of features on runtime, as our lower bounding mechanism would trivially prune these features. We instead refer to the computational complexity of our algorithm from Proposition \ref{proposition:complexityDepth2} and the number of possible decision trees as an indicative measure of the influence of the number of binary features and sparsity of the feature vectors on the runtime. 


\input{table_scalability}

\subsubsection{Section Summary}

The experimental results confirmed the efficiency of our incremental frequency computation and similarity-based lower bounding approach. Each of the techniques provides a reduction in terms of runtime. We show that dataset-based caching exhibits equal or better performance than branch-based caching across the datasets. Our approach scales approximately linearly with respect to the number of instances, and the depth of the tree has a large influence on the runtime, i.e., decision trees of depth three and four are typically computed within seconds or minutes, respectively, but trees of depth five are notably more challenging depending on the dataset. Increasing the number of binary features increases the expected runtime, but this is difficult to measure as it depends on the effectiveness of the pruning techniques for the dataset at hand. We found that inspecting features in the order as given in the dataset was more effective than ordering features according to their corresponding Gini coefficients, possibly due to the in order feature selection synergies well with incremental frequency and similarity lower bound computation, but the difference largely depends on the dataset. Lastly, our dynamic node selection strategy offered consistent improvements over a static strategy.

\subsection{Comparison Against State-Of-The-Art Optimal Decision Tree Algorithms}

Amongst the optimal decision tree methods discussed in Section \ref{section:literature}, we consider DL8.5 by \cite{dl85} as the main competing method. The rationale is that DL8.5 has been shown to largely outperform the other techniques based on generic optimisation modelling, such as integer programming (\cite{verwer2019learning,bertsimas2017optimal}) and constraint programming (\cite{verhaeghe2019learning}), when minimising the misclassification score for full binary trees. We now discuss other approaches.

The SAT method by \cite{narodytska2018learning} takes a different approach: rather than directly minimising the misclassifications given a fixed depth, it attempts to construct the smallest decision in terms of the total number of nodes that \emph{perfectly} fits the data, i.e., trees that have a misclassification score of zero. As finding the zero-misclassification tree using the complete dataset was computationally infeasible for SAT, and also prone to overfitting, the authors proposed to subsample datasets by selecting 5-20\% of the instances. While this setting has its merits, it diverges from the goals of our paper. Furthermore, we found that our algorithm computes the perfect decision tree within seconds on exactly the same subsampled data used in the SAT paper and as can be seen in tables, we can directly optimise with the complete datasets. 

Other SAT works (\cite{avellanedaefficient,janota2020sat,LNSSAT}) use either the discussed SAT method or BinOpt (\cite{verwer2019learning}) as a ground for comparison, but these have been shown to be outperformed other recent works, e.g., DL8.5, which we further improve upon. The same reasoning holds when comparing to other generic (optimisation) frameworks such as integer programming. The reason for the discrepancy in runtime between our approach and SAT is that we provide a highly specialised procedure that exploits classification tree properties, e.g., Properties \ref{property:independence} and \ref{property:overlap}. One could argue that declarative approaches are easily extendable with new constraints beyond those considered here, which may be of interest, but we do not make use of such functionality. For these reasons, we perform no further comparison with these methods.

\cite{hu2019optimal} (OSDT) and \cite{sparseICML} (GOSDT) introduce exhaustive search algorithms optimising the linear combination of the misclassification score and number of nodes in tree (Eq. \ref{eq:sparseobjective}. The pruning mechanism of both works is based on the sparsity coefficient $\alpha$ from  Eq. \ref{eq:sparseobjective}, i.e., the lower bound for each new introduce node is at least $\alpha$. As such, the sparsity coefficient $\alpha$ plays a key role in the algorithm: the larger the $\alpha$, the faster the algorithms may exhaustively explore the search space. Recall that our approach optimises the sparse objective using Algorithm \ref{alg:SparseObjective}. We experimented with both approaches, and rather than providing detailed tables, we summarise our findings. 

A direct comparison can be made with OSDT using a ten minute timeout with depth four trees. We found that our algorithm computes optimal trees with the specified objective within seconds for the benchmarks used by \cite{hu2019optimal}, whereas their method may require minutes or timeout. On the majority of our benchmarks, the approach by \cite{hu2019optimal} timeouts, unless the sparsity coefficient is set to be sufficiently high. As presented, our approach may handle any sparsity coefficient (previous tables may be interpreted as using $\alpha = 0$) within the time limit for trees with maximum depth four. Note that optimising with larger sparsity coefficient values reduces the runtime due to pruning.

The comparison with GOSDT is slightly different, since the problem definition is not the same. In GOSDT, the algorithm does not directly support limiting the depth or number of nodes, but instead the structure of the tree is controlled through the objective function and the sparsity coefficient. To facilitate a fair comparison, we first ran GOSDT on 68 datasets (all but the 'reduced' benchmarks) to observe the results. We used a time limit of ten minutes and set the sparsity coefficient to the default value\footnote{To avoid confusion, we note that our definition of the sparse objective is based on misclassifications, whereas in GOSDT it is based on accuracy. Both definitions are equivalent.} of $0.01 \cdot |D|$. We observed that GOSDT timed out on 65\% of the 68 datasets. Even though the depth is not limited, on 90\% the (possibly suboptimal) computed tree was of depth at most four, the maximum depth was seven, and all trees had a small number of nodes. These results are expected since the goal of the authors of GOSDT was to produce small trees. That said, our MurTree approach can produce optimal small trees (e.g., $depth \in \{1, 2, 3, 4\}$ and any number of nodes) within seconds or minutes even when the sparsity coefficient is set to zero. Note that after running our algorithm using value zero for the sparsity coefficient, the cache will be populated, and then a tree for any sparsity value may be extracted immediately. Intuitively, the zero-case coefficient is the worst case, and optimising with greater values of the sparsity coefficient is beneficial as it offers pruning.

We now proceed with the main comparison with DL8.5.


\subsubsection{Comparison with DL8.5 by \cite{dl85}}

The aim is to assess the effectiveness of our MurTree approach with respect to DL8.5, the state-of-the-art method for optimal decision trees. We evaluate the runtime of both methods to exhaustively explore the search space: a lower runtime indicates a more effective approach.

DL8.5 optimises the misclassification score given a constraint on the depth of the tree. The number of feature nodes cannot be limited, meaning that full binary trees are considered, i.e., effectively the number of feature nodes is set to the maximum value given the depth. We experiment with maximum depths of four and five. To ensure a fair comparison, i.e., have both algorithms solve exactly the same problem, we set the maximum number of feature nodes for our method to 15 for depth four and 31 for depth five trees. The complete dataset is given to both algorithms without dividing into the training and test set. Ten minutes is allocated for each dataset.

Although it is standard practice in machine learning to compare learning algorithms on out-of-sample accuracy, in this case runtime is more appropriate for evaluating the methods since both algorithms are optimising the same objective. The out-of-sample accuracy evaluation of optimal decision trees is reserved for the next section. Note that since the algorithms discriminate trees solely based on the objective, the resulting trees obtained by both methods, assuming neither method timed out, will have the same objective value but may differ in their structure and features selected.

The runtimes, given in Table \ref{table:vsdl85}, show that our method is orders of magnitude faster than DL8.5. This is a significant result, as DL8.5 has been previously shown to outperform other techniques for optimal classification trees based on integer and constraint programming by a large margin. Our results illustrate the advantage of designing and specialising decision tree optimisation algorithms compared to using off-the-shelf tools. Both DL8.5 and our MurTree approach exploit the dynamic programming structure of decision trees, but our method employs additional techniques to further take advantage of the properties of decision trees. The reduced runtime contributes greatly towards the application of optimal classification tree methods in practice, especially when tuning is involved (see next section).


\input{comparison_vs_dl85}

\subsection{Comparison Against Conventional Algorithms on Out-Of-Sample Accuracy}
\label{section:outofsample}

In this section, we analyse the suitability of our optimal decision trees as out-of-sample classifiers. The aim is to demonstrate that more accurate trees of limited size lead to better generalisations than what is offered by heuristic approaches. Note that the restricted size of optimal decision trees plays the role of a regulariser to avoid overfitting. The main comparison is done against an optimised implementation of CART (\cite{breiman1984classification}), a widely used decision tree learning algorithm. For illustrative purposes, we also make a comparison with random forests, as a related method that typically improves accuracy over standard decision tree algorithms at the expense of being less interpretable. As will be discussed, our experiments further confirm similar empirical findings (\cite{verwer2019learning,bertsimas2017optimal}). The experiments were run on exactly the same (binarised) benchmarks for all methods. We use the algorithms provided by the \emph{sklearn} (\cite{scikit-learn}) Python package for machine learning for the other methods.

\subsubsection{Hyper-Parameter Tuning}

Selecting a good set of parameters is important when evaluating the performance of machine learning models. Hyper-parameter tuning is performed for all methods using grid search. Given a model, we compute the average train and test accuracy using stratified five-fold cross-validation for each combination of parameters. The set of parameters that leads to the best test accuracy is selected. Note that the model is trained on the training sets, but evaluated on the test sets. The runtime presented includes the time taken to perform cross-fold validation using \emph{all} parameters and the time to train a new decision using the selected parameters on the complete dataset. All methods and parameter configurations used exactly the same folds.

\subsubsection{Comparison Against Heuristic Decision Trees (CART)}

We considered three tuning settings for our MurTree method to analyse the effect of restricting tuning options. The three settings are as follows:

\begin{enumerate}
\item MT-F: Only a single parameter configuration is set based on a heuristically obtained tree. The parameter values are fixed to match those produced by the decision tree computed using CART. Note that, strictly speaking, this is not a hyper-tuning approach, but nevertheless gives insight on the generalisability of optimal decision trees.
\item MT-R: The heuristically obtained decision tree provides an upper bound on the allowed parameter values for tuning, i.e., given a tree constructed by CART with depth $d$ and number of feature nodes $s$, tuning is done with $depth \in {1, ..., d}$ and $feature\ node\ count \in \{depth, ..., s\}$.
\item MT-A: Fully exploit available parameters of our algorithm until depth four, i.e., $depth \in \{1, 2, 3, 4\}$ and $num\_feature\_nodes \in \{depth, depth+1, ..., 2^{depth}-1\}$.
\end{enumerate}

\input{figure_out_of_sample_accuracy}

The aggregated results on 84 datasets are shown in Figure \ref{figure:outofsample} for the three settings compared to CART, which was tuned using $depth \in [1, 2, 3, 4]$.

When considering the MT-F parameter selection strategy, the results are roughly comparable to the outcome produced by CART, even though optimal decision trees have been training accuracy. The mismatch between better training and lack of consistent performance on the test set indicates that the structure of the tree produced by CART may be suboptimal for the considered dataset.

The performance is notably different when allowing more freedom in parameter selection. The MT-R parameter selection strategy produces better results for most datasets, while taking into account all parameter options (MT-A) consistently demonstrates greater out-of-sample accuracy across the datasets.

The runtime of our MurTree approach is reasonably short for most benchmarks. However, as expected, CART is  much faster, i.e., the runtime was only a fraction of a second for almost all benchmarks. Nevertheless, the runtime difference between the methods may be acceptable for a vast number of application, especially when training time is not a concern.

Overall, we conclude that the trees produced by our MurTree algorithm provide better generalisation compared to trees obtained using CART, a classification decision tree algorithm, at the expense of greater runtime.

\subsubsection{Comparison Against Random Forests}

For completeness we show a comparison with tuned random forests using the same sklearn Python package. A forest of trees is typically more accurate than a single decision tree, but the resulting model is less concise and more difficult for human interpretation. The forests were tuned by varying the number of trees in the forest from $[10, 50, 100]$, selecting the maximum depth from $[no\_limit, 1, 2, 4]$, and considering a subset of the features at each step to evaluate with respect to $[|\mathcal{F}|, \frac{1}{2}|\mathcal{F}|, \sqrt{|\mathcal{F}|}, log_2(|\mathcal{F}|)]$, where $\mathcal{F}$ is the set of features.

We show the difference in both training and test accuracy in Figure \ref{figure:vsForests}. It is evident that random forests have the advantage on the training set, which translates to the test set. Nevertheless, for roughly half of the datasets optimal trees of depth four achieve comparable performance in terms of accuracy. This demonstrates that for some applications optimal decision trees may be preferred over heuristically trained random forests. The runtime of random forest is not negligible but still reasonable: majority of the datasets fall into the 10-19 seconds range. We note that a different tuning strategy for random forests may lead to lower/higher runtimes.

\input{figure_out_of_sample_vs_forest}

\section{Conclusion}
\label{section:conclusion}

We presented MurTree, an algorithm for computing optimal decision trees, i.e., decision trees that achieve the best representation of the data in terms of the misclassification score. The algorithm is based on dynamic programming and search. Our novel techniques exploit decision tree properties to provide orders of magnitude speed-ups when compared to the state-of-the-art. The conducted experimental study shows that optimal decision trees are desirable as their out-of-sample accuracy is greater than decision trees obtained using a conventional learning algorithm (CART), while providing concise and interpretable models within reasonable time for the majority of the benchmarks.

There are several limitations of our algorithm, some of which are shared with other optimal decision tree algorithms. The depth of the trees is kept relatively low, e.g., depth four. A low depth is convenient for interpretability, but for some applications, deeper trees may be necessary, e.g., compactly representing controllers using perfect trees (\cite{ashok2020dtcontrol}). We observed that for half of the datasets considered, optimal decision trees provide comparable performance in terms of out-of-sample accuracy when compared to the more complex model of random forests, but for the other half of datasets, random forests had better generalisation. In the setting we consider, the predicates are required to be provided in advance, i.e., the dataset must be binarised. Given that the algorithm is unlikely to support tens of thousands of predicates in the current form, a trade-off must be made between the runtime and number of predicates when using datasets with continuous or categorical features. Even though our algorithm provides significant speed-ups, traditional heuristic methods remain much faster. Nevertheless, for most tested datasets our approach produced optimal trees within seconds or minutes, which may be acceptable for offline applications. 

There are several directions for future work. Considering novel metrics to improve the ability to generalise better on unseen data may be one such direction, or understand which optimal tree to select out of a set of trees with minimum misclassification scores. Analysing the effect of supervised discretisation algorithms for binarising the datasets may lead to additional insights. Furthermore, constructing forests of optimal trees is another research direction worth considering.



\acks{We would like to acknowledge the comments of the editor and the anonymous reviewers. Their commitment to the reviewing process has considerably contributed towards clarity, accessibility, and correctness of the paper. An anonymous reviewer motivated us to explore caching based on datasets which led to improvements. Part of this work was done while Anna Lukina was visiting the Simons Institute for the Theory of Computing.}






\vskip 0.2in
\bibliography{references}

\end{document}

%% file: algo_main.tex
\begin{algorithm}
\small
		\DontPrintSemicolon
		\SetKwInput{Input}{input}
		\SetKwInput{Output}{output}
		\Input{Dataset $\mathcal{D} = \mathcal{D^+} \cup \mathcal{D^-}$, branch of the subtree $B$, depth $d \in \mathbb{N}_0$, number of feature nodes $n \in \mathbb{N}_0$, upper bound on the misclassifications $UB \in \mathbb{Z}$}
		\Output{Optimal classification tree within the input specification that minimises the misclassification score on $\mathcal{D}$}
		\Begin{
	
		\tcp{Prune based on the upper bound}
		\If{$UB < 0\label{alg1:line:UBtestSTART}$}
		{
				\textbf{return} $\emptyset$ \label{alg1:line:UBtestEND}\;		
		}

		\tcp{Base case, second case (Eq. \ref{DPformulation}): no feature nodes are possible}
		\If{$d = 0 \lor n = 0\label{alg1:line:basecaseSTART}$}
		{
			\If{$LeafMisclassification(\mathcal{D}) \leq UB$}
			{
				\textbf{return} $ClassificationNode(\mathcal{D})$\;
			}
			\Else
			{
				\textbf{return} $\emptyset$\label{alg1:line:basecaseEND}\;	
			}							
		}	
		\tcp{Use cached subtrees if possible (Section \ref{section:caching})}
		\If{IsOptimalSubtreeInCache$(\mathcal{D}, B, d, n)\label{alg1:line:cacheTestSTART}$}
		{
			$T \leftarrow RetrieveOptimalSubtreeFromCache(\mathcal{D}, B, d, n)$\;
			\If{$Misclassifications(T) \leq UB$}
			{
				\textbf{return} $T$\;
			}
			\Else
			{
				\textbf{return} $\emptyset$\label{alg1:line:cacheTestEND}\;	
			}	
		}

		\tcp{Update the cache using the similarity-based lower bound (Section \ref{section:similarityBounding})}
		\tcp{Note that an optimal solution may be found in the process}
		$updated\_optimal\_solution \leftarrow UpdateCacheUsingSimilarity(\mathcal{D}, B, d, n)$\;
		\If{$updated\_optimal\_solution\label{alg1:line:cacheUpdateSTART}$}
		{
			$T \leftarrow RetrieveOptimalSubtreeFromCache(\mathcal{D}, B, d, n)$\;
			\If{$Misclassifications(T) \leq UB$}
			{
				\textbf{return} $T$\;
			}
			\Else
			{
				\textbf{return} $\emptyset$\label{alg1:line:cacheUpdateEND}\;	
			}
		}

		\tcp{Prune if the lower bound exceeds the upper bound, since no tree can be found within the upper bound requirement (Section \ref{section:LBcache})}
		$LB \leftarrow RetrieveLowerBoundFromCache(\mathcal{D}, B, d, n)$\label{alg1:line:LBPruneSTART}\;
		\If{$LB > UB$}
		{
			\textbf{return} $\emptyset$\;
		}	

		\tcp{If the leaf node is already at the lower bound, no need to look further}
		\If{$LB = LeafMisclassification(\mathcal{D})$}
		{
			\textbf{return} $ClassificationNode(\mathcal{D})\label{alg1:line:LBPruneEND}$\;
		}
					
		\tcp{Use Algorithm \ref{algorithm:specialisedDT} for small trees from Section \ref{section:specialisedAlgorithm}}
		\tcp{Note that the specialised algorithm updates the cache}
		\If{$d \leq 2\label{alg1:line:specialisedAlgSTART}$}
		{
			$T \leftarrow SpecialisedDepthTwoAlgorithm(\mathcal{D}, B, d, n)$\;
			\If{$Misclassifications(T) \leq UB$}
			{
				\textbf{return} $T$\;
			}
			\Else
			{
				\textbf{return} $\emptyset\label{alg1:line:specialisedAlgEND}$\;	
			}				
		}
		
		\tcp{Fourth case (Eq. \ref{DPformulation}): exhaustively search using Algorithm \ref{main_algo_general_case}}
		\textbf{return} $MurTree.GeneralCase(\mathcal{D}, B, d, n, UB)$\label{alg1:line:recursion}\;		
		}
		\caption{$MurTree.SolveSubtree(\mathcal{D}, B, d, n, UB)$, the main algorithm loop\label{algo_main}}
	\end{algorithm}

%% file: algo_main_general_case.tex
\begin{algorithm}
		\DontPrintSemicolon
		\small
		\DontPrintSemicolon
		\SetKwInput{Input}{input}
		\SetKwInput{Output}{output}
		\Input{Dataset $\mathcal{D} = \mathcal{D^+} \cup \mathcal{D^-}$, branch of the subtree $B$, depth $d \in \mathbb{N}_0$, number of feature nodes $n \in \mathbb{N}_0$, upper bound on the misclassifications $UB \in \mathbb{Z}$}
		\Output{Optimal classification tree within the input specification that minimises the misclassification score on $\mathcal{D}$. }

		\Begin{
		\tcp{Use a single classification node as an initial solution}
		$T_{best} \leftarrow ClassificationNode(\mathcal{D})$\;
		\If{$LeafMisclassification(\mathcal{D}) > UB$}
		{
			$T_{best} \leftarrow\emptyset$\;
		} 
	    \tcp{Find the lower bound stored in cache (Section \ref{section:LBcache})}
		$LB \leftarrow RetrieveLowerBoundFromCache(\mathcal{D}, B, d, n)$\;
		\tcp{RLB refers to the refined lower bound in Eq. \ref{eq:slb}}	
		$RLB \leftarrow \infty$\;
		\tcp{Compute allowed number of nodes for child subtrees}	
		$n_{max} \leftarrow \min\{2^{(d-1)}-1, n-1\}$\;
		$n_{min} \leftarrow  (n - 1 - n_{max})$\;		
		\For{$splitting\ feature\ f \in \mathcal{F}\label{alg2:line:split}$}
		{
			\tcp{If the current best node is the optimal node, stop}	
			\If{$Misclassifications(T_{best}) = LB\label{alg2:line:LBSTART}$}
			{
				$\textbf{break}\label{alg2:line:LBEND}$
			}

			\tcp{Nondiscriminary splits should be avoided}
			\If{$|\mathcal{D}(\overline{f})| = 0 \lor |\mathcal{D}(f)| = 0\label{alg2:line:discriSTART}$}
			{
				$\textbf{continue}\label{alg2:line:discriEND}$
			}	

			\For{$n_{L} \in [n_{min}, n_{max}]$\label{alg2:line:nodeBudgetSTART}}
			{
				$n_{R} \leftarrow n - 1 - n_{L}$\;
				\tcp{Impose an upper bound $UB'$ that ensures that a feasible tree will have fewer misclassifications than the best tree found so far $T_{best}$}
				$UB' \leftarrow \min\{UB, Misclassifications(T_{best})-1\}$\;
				\tcp{Use Algorithm \ref{main_algo_general_case_helper} to compute subproblem}
				$(T, LB_{local}) \leftarrow MurTree.SolveSubtreeGivenRootFeature(\mathcal{D}, B, f, d, n_{L}, n_{R}, UB')$\;
				\If{$T \neq \emptyset$}
				{
					$T_{best} \leftarrow T\label{alg2:line:bestUpdate}$\;
				}
				\Else
				{
					$RLB \leftarrow \min\{RLB, LB_{local}\}\label{alg2:line:nodeBudgetEND}$\;		
				}				
			}				
		}		
		\tcp{Cache the optimal solution...}
		\If{$Misclassifications(T_{best}) \leq UB\label{alg2:line:cacheUpdatesSTART}$}
		{
			$StoreOptimalSubtreeInCache(T_{best}, \mathcal{D}, B, d, n)$\;
		}
		\tcp{...or record the lower bound (Section \ref{section:LBstore})}
		\Else
		{
			$LB \leftarrow \max\{LB, UB+1\}$\;
			\If{$RLB < \infty$}
			{
				$LB \leftarrow \max\{LB, RLB\}$\;
			}
			\tcp{Store the lower bound in the cache (Section \ref{section:LBstore})}
			$StoreLowerBoundInCache(\mathcal{D}, B, d, n, LB)\label{alg2:line:cacheUpdatesEND}$\;		
		}
		$ReplaceDatasetForSimilarityBound(\mathcal{D}, B, d)$\;
		\textbf{return} $T_{best}$\;					    
		}
		\caption{$MurTree.GeneralCase(\mathcal{D}, B, d, n, UB)$, the general (fourth) case of Eq. \ref{DPformulation} used in Algorithm \ref{algo_main} \label{main_algo_general_case}}
	\end{algorithm}

%% file: algo_main_general_case_2.tex
\begin{algorithm}[t]
		\DontPrintSemicolon
		\DontPrintSemicolon
		\SetKwInput{Input}{input}
		\SetKwInput{Output}{output}
		\Input{Dataset $\mathcal{D} = \mathcal{D^+} \cup \mathcal{D^-}$, branch of the subtree $B$, root feature $f_{root} \in \mathcal{F}$, depth $d \in \mathbb{N}_0$, number of feature nodes in left and right subtree $n_L, n_R \in \mathbb{N}_0$, upper bound on the misclassifications $UB \in \mathbb{Z}$}
		\Output{An optimal decision tree with feature $f_{root}$ as its root that satisfies the input specification and minimises the misclassification score on $\mathcal{D}$. If no such tree exists, a lower bound on the misclassification score is returned.}

		\Begin{
		\tcp{Get the depth and branches of the children subtrees}	
		$d_L \leftarrow \min(d-1, n_L)$\;								
		$d_R \leftarrow \min(d-1, n_R)$\;	
		$(B_L, B_R) \leftarrow GetChildBranches(B, f_{root})$\;						
		\tcp{Dynamic order: process left subtree first (Section \ref{section:nodeselection})}					
					\If{$LeafMisclassification(\mathcal{D}(\overline{f}_{root})) > LeafMisclassification(\mathcal{D}(f_{root}))$}
					{
					$UB_L \leftarrow UB - RetrieveLowerBoundFromCache(\mathcal{D}(\overline{f}_{root}), B_R, d_R, n_R) $\;
					
$T_L \leftarrow MurTree(\mathcal{D}(\overline{f}_{root}), B_L, d_L, n_L, UB_L)$

					\tcp{No need to compute the right subtree if the left child is infeasible}
					\If{$T_L = \emptyset$}
					{
							$LB_{local} \leftarrow\ $compute local bound (Eq. $\ref{eq:localBound}$)\;
							$\textbf{return}\ (\emptyset, LB_{local})$
					}		

	$UB_R \leftarrow UB - Misclassifications(T_L)$\;		

					$T_R \leftarrow MurTree(\mathcal{D}(f_{root}), B_R, d_R, n_R, UB_R)$

					\tcp{If both children are feasible, return the optimal solution}
					\If{$T_R \neq \emptyset$}
					{
						$T \leftarrow\ $tree with root feature $f_{root}$ and children $T_L$ and $T_R$\;
						$\textbf{return}\ (T, Misclassifications(T))$
					}
					\Else
					{
			$LB_{local} \leftarrow\ $compute local bound (Eq. $\ref{eq:localBound}$)\;
			$\textbf{return}\ (\emptyset, LB_{local})$	
					}				
				}
				\Else
				{
				\tcp{Dynamic post-order: process right subtree first (Section \ref{section:nodeselection})}
				Process right subtree analogously as above\;
				}
				    
		}
		\caption{$MurTree.SolveSubtreeGivenRootFeature(\mathcal{D}, B, f_{root}, d, n_{L}, n_{R}, UB)$: a subroutine used as part of Algorithm \ref{main_algo_general_case} \label{main_algo_general_case_helper}}
	\end{algorithm}

%% file: algo_specialised.tex
	\begin{algorithm}[t]
		\DontPrintSemicolon
		\SetKwInput{Input}{input}
		\SetKwInput{Output}{output}
		\SetKwFunction{Update}{Update}
		\Input{Binary dataset $\mathcal{D} = \mathcal{D^+} \cup \mathcal{D^-}$}
		\Output{Optimal classification tree of depth two with three feature nodes that minimises the misclassification score on $\mathcal{D}$ }
		\Begin{
				$\forall f_i: FQ^+(f_i) \leftarrow 0 \land FQ^-(f_i) \leftarrow 0\label{specialisedtree:PhaseOneStart}$\;
				$\forall f_i, f_j, i<j: FQ^+(f_i, f_j) \leftarrow 0 \land FQ^-(f_i, f_j) \leftarrow 0$\;
				\tcc*{Step 1: construct the frequency counter of positive features}
		    \For{$\mathbf{fv} \in \mathcal{D^+}$}
		    {
		    		\For{$f_i \in \mathbf{fv}$}
		    		{
		    				$increment\ FQ^+(f_i)$\;
		    				\For{$f_j \in \mathbf{fv}\ s.t.\ i<j$}
		    				{
		    						$increment\ FQ^+(f_i, f_j)$\;
			    				}	 
		    		}	 
		    }	 
		    $FQ^-$ is computed as above using $\mathcal{D^-}\label{specialisedtree:PhaseOneEnd}$\;   		    
			\tcc*{Step 2: construct the optimal decision tree based on the frequency counters $FQ^+$ and $FQ^-$}
			\tcc*{Compute the best left and right subtrees for each feature} 
			\For{$f_i \in \mathcal{F}\label{specialisedtree:PhaseTwoStart}$}
		  	{
		  	\For{$f_j \in \mathcal{F}\ s.t.\ i \neq j$}
		  	{
		  			$CS(\overline{f_i}, f_j) \leftarrow \min\{FQ^+(\overline{f_i}, f_j), FQ^-(\overline{f_i}, f_j)\}$\;
		  			$CS(\overline{f_i}, \overline{f_j}) \leftarrow \min\{FQ^+(\overline{f_i}, \overline{f_j}), FQ^-(\overline{f_i}, \overline{f_j})\}$\;
		  			\tcc*{Compute branch with $f_i$ as root and $f_j$ as left child}
		  			$MS_{left}(f_i, f_j) \leftarrow CS(\overline{f_i}, \overline{f_j}) + CS(\overline{f_i}, f_j)$\;
					\If{$BestLeftSubtree(f_i).misclassification > MS_{left}(f_i, f_j)$}
					{
		  				$BestLeftSubtree(f_i).misclassification \leftarrow MS_{left}(f_i, f_j)$\;
		  				$BestLeftSubtree(f_i).feature \leftarrow f_j$\;
		  			}
		  			The best right subtree with $f_i$ as the root and $f_j$ as the right child is computed analogously as above\; 
		  	}		   
			}	 
			\tcc*{Compute the best tree by taking the feature with the minimum sum of misclassification of its children}
			$best\_tree \leftarrow argmin_{f_i \in \mathcal{F}}\{BestLeftSubtree(f_i).misclassification + BestRightSubtree(f_i).misclassification\}\label{specialisedtree:PhaseTwoEnd}$\;
		   $\KwRet\ best\_tree$\;
		}
		\caption{Specialised algorithm for computing optimal classification trees of depth two with three nodes \label{algorithm:specialisedDT}}
\end{algorithm}

%% file: algo_hashing.tex
\begin{algorithm}
\small
		\DontPrintSemicolon
		\SetKwInput{Input}{input}
		\SetKwInput{Output}{output}
		\Input{Array of integers $A = [a_1, a_2, ..., a_n]$}
		\Output{An integer $k$ representing the hash value of $A$}
		\Begin{
		$k \leftarrow n$\;
		\For{$i \in [1, n]$}
		{
			$k \leftarrow k \oplus (a_i + 0x9e3779b9 + 64k + k/4)$\;			
		}
		\textbf{return} k\;
		}
		\caption{A standard hash function for an array of integers \label{alg:hash}}
	\end{algorithm}

%% file: algo_sparse_objective.tex
\begin{algorithm}
\small
		\DontPrintSemicolon
		\SetKwInput{Input}{input}
		\SetKwInput{Output}{output}
		\Input{Dataset $\mathcal{D} = \mathcal{D^+} \cup \mathcal{D^-}$, depth $d \in \mathbb{N}_0$, number of feature nodes $n \in \mathbb{N}_0$, sparse coefficient $\alpha \in \mathbb{N}_0$}
		\Output{Optimal decision tree within the input specification that minimises the sparse objective (Eq. \ref{eq:sparseobjective}) on dataset $\mathcal{D}$}
		\Begin{

		$T_{best} \leftarrow ClassificationNode(\mathcal{D})$\;
		\For{$n' \in [1, n]$}
		{
			$UB \leftarrow SparseObjective(T_{best}) - (\alpha \cdot n') - 1$\;
			$T \leftarrow MurTree.SolveSubtree(\mathcal{D}, \emptyset, d, n', UB)$\;
			\If{$T \neq \emptyset$}
			{
				$T_{best} \leftarrow T$\;
			}
		}
		\textbf{return} $T_{best}$\;
		}
		\caption{$MurTree.SolveSparseObjective(\mathcal{D}, d, n, \alpha)$, an algorithm to minimise the sparse objective (Eq.\ref{eq:sparseobjective}) \label{alg:SparseObjective}}
	\end{algorithm}

%% file: algo_minimising_num_nodes.tex
\begin{algorithm}
\small
		\DontPrintSemicolon
		\SetKwInput{Input}{input}
		\SetKwInput{Output}{output}
		\Input{Dataset $\mathcal{D} = \mathcal{D^+} \cup \mathcal{D^-}$, depth $d \in \mathbb{N}_0$, number of feature nodes $n \in \mathbb{N}_0$}
		\Output{Optimal decision tree that lexicographically minimise the misclassification and then the number of nodes on dataset $\mathcal{D}$}
		\Begin{
		$T_{best} \leftarrow MurTree.SolveSubtree(\mathcal{D}, \emptyset, d, n, UB)$\;
		$UB \leftarrow Misclassification(T_{best})$\;
		\For{$n' \in [n-1, 0]$}
		{			
			$T \leftarrow MurTree.SolveSubtree(\mathcal{D}, \emptyset, d, n', UB)$\;
			\If{$T \neq \emptyset$}
			{
				$T_{best} \leftarrow T$\;
			}
		}
		\textbf{return} $T_{best}$\;
		}
		\caption{$MurTree.SolveSubtreeLexicographically(\mathcal{D}, d, n)$, an algorithm to compute the tree with minimum misclassifications using the least amount of nodes\label{alg:minNodes}}
	\end{algorithm}

%% file: table_hash.tex
\begin{table}
\caption{Comparing the efficiency of branch and dataset caching. For each dataset, the number of instances ($\mathcal{D}$), binary features ($\mathcal{F}$), and number of classes ($\mathcal{C}$) are displayed. Datasets where the difference between the methods is insignificant are excluded. The time represents the number of seconds to compute decision trees with $n \in [1, 15]$ feature nodes with maximum depth four (fifteen trees in total). The number of cache entries is given in thousands. Bold numbers highlight better runtime.\label{table:hashComparison}}%
    \centering
    \scriptsize{
\def\d{@{\hspace*{1.7mm}}}
\begin{tabular}{\d l | c | c | c | c | c | c | c | \d}
& & & & \multicolumn{2}{c | }{Branch Cache} & \multicolumn{2}{c}{Dataset Cache} \\ \hline
Name & $|\mathcal{D}|$ & $|\mathcal{F}|$ & $|\mathcal{C}|$	& Time & Cache Entries & Time & Cache Entries \\ \hline
ionosphere & 351 & 445 & 2	& 250 & 170 & \textbf{132} & 54 \\
letter & 20000 & 224 & 2	& 417 & 32 & \textbf{264} & 11 \\
pendigits & 7494 & 216 & 2	& 153 & 34 & \textbf{88} & 11 \\
segment & 2310 & 235 & 2	& 11 & 12 & \textbf{8} & 4 \\
splice-1 & 3190 & 287 & 2	& 111 & 44 & \textbf{99} & 35 \\
vehicle & 846 & 252 & 2	& 31 & 40 & \textbf{18} & 13 \\
Statlog\_satellite & 4435 & 385 & 6	& \textbf{507} & 97 & 519 & 95 \\
Statlog\_shuttle & 43500 & 181 & 7	& \textbf{114} & 17 & 124 & 17 \\
appendicitis & 106 & 530 & 2	& 12 & 140 & \textbf{7} & 22 \\
australian & 690 & 1163 & 2	& 1001 & 699 & \textbf{740} & 463 \\
backache & 180 & 475 & 2	& 48 & 120 & \textbf{34} & 72 \\
cleve & 303 & 395 & 2	& 14 & 83 & \textbf{12} & 73 \\
colic & 368 & 415 & 2	& 55 & 105 & \textbf{41} & 86 \\
heart-statlog & 270 & 381 & 2	& 12 & 77 & \textbf{9} & 66 \\
hepatitis & 155 & 361 & 2	& 19 & 72 & \textbf{12} & 41 \\
hungarian & 294 & 330 & 2	& 16 & 60 & \textbf{12} & 50 \\
new-throid & 215 & 334 & 3	& 35 & 56 & \textbf{32} & 39 \\
promoters & 106 & 334 & 2	& 32 & 61 & \textbf{27} & 54
\end{tabular}
}
\end{table}

%% file: table_our_techniques.tex
\begin{table}[t]
\caption{Comparison to determine effectiveness of the incremental frequency counter computation (`Inc') and the similarity-based lower bound (`SLB'). For each dataset, the number of instances ($\mathcal{D}$), number of binary features ($\mathcal{F}$), and number of classes ($\mathcal{C}$) are displayed. Datasets where the effect of similarity-based lower bounding is insignificant are excluded. The time represents the number of seconds the algorithms require to compute decision trees with $n \in [1, 15]$ feature nodes with maximum depth four (fifteen trees in total).  Timeouts (1800 seconds) denoted as ---. Bold numbers highlight better runtime.\label{table:partZwei}}%
    \centering
    \scriptsize{
\def\d{@{\hspace*{1.7mm}}}
\begin{tabular}{\d l | c | c | c | r r r \d}
Name & $|\mathcal{D}|$ & $|\mathcal{F}|$ & $|\mathcal{C}|$ & noInc-noSLB & inc-noSLB & inc-SLB \\ \hline
letter & 20000 & 224 & 2 & 810 & \textbf{237} & 269 \\
pendigits & 7494 & 216 & 2 & 185 & \textbf{86} & 96 \\
segment & 2310 & 235 & 2 & 49 & 15 & \textbf{9} \\
default\_credit & 30000 & 44 & 4 & 7 & \textbf{7} & 9 \\
magic04 & 19020 & 79 & 2 & 8 & \textbf{6} & 9 \\
Statlog\_satellite & 4435 & 385 & 6 & 1060 & 756 & \textbf{516} \\
Statlog\_shuttle & 43500 & 181 & 7 & 131 & \textbf{88} & 128 \\
appendicitis & 106 & 530 & 2 & 32 & 28 & \textbf{7} \\
australian & 690 & 1163 & 2 & --- & --- & \textbf{788} \\
backache & 180 & 475 & 2 & 86 & 77 & \textbf{37} \\
cleve & 303 & 395 & 2 & 65 & 58 & \textbf{14} \\
colic & 368 & 415 & 2 & 89 & 73 & \textbf{43} \\
heart-statlog & 270 & 381 & 2 & 55 & 49 & \textbf{10} \\
hepatitis & 155 & 361 & 2 & 30 & 26 & \textbf{13} \\
hungarian & 294 & 330 & 2 & 31 & 28 & \textbf{15} \\
new-throid & 215 & 334 & 3 & 148 & 143 & \textbf{35} \\
shuttleM & 14500 & 691 & 2 & 1062 & 805 & \textbf{416}
\end{tabular}
}
\end{table}

%% file: table_feature_node.tex
\begin{table}
\caption{Comparison of feature and node selection strategies. For each dataset, the number of instances ($\mathcal{D}$), number of binary features ($\mathcal{F}$), and number of classes ($\mathcal{C}$) are displayed. The time represents the number of seconds the algorithms require to compute decision trees with $n \in [1, 15]$ feature nodes with maximum depth four (fifteen trees in total). Datasets solved under a second or when the differences between the variants was negligible have been omitted.\label{table:partthree}}%
    \centering
    \scriptsize{
\def\d{@{\hspace*{1.7mm}}}
\begin{tabular}{\d l | c | c | c | c | c | c | c | c | \d}
\multicolumn{2}{l |}{Feature Selection} & \multicolumn{2}{c |}{} & Gini & Random & \multicolumn{2}{c |}{InOrder} \\ \hline 
\multicolumn{2}{l |}{Node Selection} & \multicolumn{2}{c |}{} & Dynamic & Dynamic & PostOrder & Dynamic \\ \hline 
Name & $|\mathcal{D}|$ & $|\mathcal{F}|$ & $|\mathcal{C}|$	& Time & Time & Time & Time \\ \hline
ionosphere & 351 & 2 & 2& 154 & 332 & 139 & \textbf{132} \\
letter & 20000 & 2 & 2& 310 & 443 & 276 & \textbf{264} \\
pendigits & 7494 & 2 & 2& 100 & 133 & 93 & \textbf{88} \\
segment & 2310 & 2 & 2& 16 & 23 & 10 & \textbf{8} \\
splice-1 & 3190 & 2 & 2& \textbf{94} & 128 & 103 & 99 \\
vehicle & 846 & 2 & 2& 23 & 30 & 22 & \textbf{18} \\
default\_credit & 30000 & 4 & 4& 9 & 9 & \textbf{7} & \textbf{7} \\
magic04 & 19020 & 2 & 2& 11 & 11 & \textbf{9} & \textbf{9} \\
Statlog\_satellite & 4435 & 6 & 6& \textbf{458} & 665 & 518 & 519 \\
Statlog\_shuttle & 43500 & 7 & 7& 140 & 187 & 126 & \textbf{124} \\
australian & 690 & 2 & 2& \textbf{445} & 1155 & 803 & 740 \\
backache & 180 & 2 & 2& \textbf{24} & 40 & 37 & 34 \\
cleve & 303 & 2 & 2& \textbf{10} & 17 & 14 & 12 \\
colic & 368 & 2 & 2& 44 & 60 & 43 & \textbf{41} \\
heart-statlog & 270 & 2 & 2& \textbf{8} & 14 & 11 & 9 \\
hepatitis & 155 & 2 & 2& \textbf{12} & 18 & 14 & \textbf{12} \\
hungarian & 294 & 2 & 2& 13 & 17 & 14 & \textbf{12} \\
promoters & 106 & 2 & 2& \textbf{21} & 31 & 30 & 27 \\
shuttleM & 14500 & 2 & 2& 466 & 468 & 477 & \textbf{383} 
\end{tabular}
}
\end{table}

%% file: table_scalability.tex
\begin{table}[t]
\caption{Scalability of our MurTree approach as instances are duplicated two, three and four times with varying depth. For each dataset, the number of instances ($\mathcal{D}$), number of binary features ($\mathcal{F}$), and number of classes ($\mathcal{C}$) are displayed. Results for a subset of all datasets are shown for simplicity. The time represents the number of seconds the algorithms require to compute decision trees with $n \in [1, 15]$ for $depth = 4$ (fifteen trees) and $n \in [1, 31]$  for $depth = 5$ (thirty-one trees). The number of cache entries is given in thousands. \label{table:scalability}}%
    \centering
    \scriptsize{
\def\d{@{\hspace*{1.7mm}}}
\begin{tabular}{ \d l | c | c | c | c c c c | c | c c c c | c \d }
  &   &   &   & \multicolumn{4}{c |}{Time for depth=4}	&  &	\multicolumn{4}{c}{Time for depth=5}	&  \\ \hline
Name & $|\mathcal{D}|$ & $|\mathcal{F}|$ & $|\mathcal{C}|$ & $1  |\mathcal{D}|$ & $2  |\mathcal{D}|$ & $3  |\mathcal{D}|$	& $4  |\mathcal{D}|$	& cache &	$1  |\mathcal{D}|$	& $2  |\mathcal{D}|$ &	 $3  |\mathcal{D}|$ &	 $4  |\mathcal{D}|$	& cache \\ \hline
anneal & 812 & 93 & 2 & 1 & 1 & 1 & 2 & 2 & 10 & 15 & 20 & 27 & 22\\
audiology & 216 & 148 & 2 & 1 & 1 & 1 & 1 & 3 & 14 & 20 & 25 & 30 & 79\\
australian-credit & 653 & 125 & 2 & 2 & 3 & 4 & 5 & 6 & 76 & 95 & 125 & 160 & 157\\
breast-wisconsin & 683 & 120 & 2 & 1 & 2 & 2 & 3 & 3 & 17 & 26 & 37 & 47 & 56\\
diabetes & 768 & 112 & 2 & 2 & 3 & 4 & 5 & 5 & 65 & 91 & 119 & 146 & 118\\
german-credit & 1000 & 112 & 2 & 3 & 5 & 7 & 9 & 10 & 160 & 232 & 308 & 381 & 385\\
heart-cleveland & 296 & 95 & 2 & 1 & 1 & 1 & 2 & 3 & 19 & 25 & 31 & 36 & 74\\
hypothyroid & 3247 & 88 & 2 & 2 & 3 & 5 & 6 & 2 & 40 & 60 & 85 & 106 & 27\\
kr-vs-kp & 3196 & 73 & 2 & 1 & 2 & 3 & 4 & 1 & 17 & 29 & 42 & 54 & 16\\
mushroom & 8124 & 119 & 2 & 5 & 9 & 13 & 18 & 5 & 101 & 234 & 345 & 478 & 130\\
segment & 2310 & 235 & 2 & 8 & 15 & 21 & 30 & 4 & 60 & 105 & 155 & 202 & 44\\
yeast & 1484 & 89 & 2 & 1 & 3 & 3 & 4 & 3 & 35 & 61 & 88 & 108 & 57\\
biodeg & 1055 & 81 & 2 & 2 & 2 & 4 & 4 & 8 & 75 & 104 & 135 & 161 & 261\\
default\_credit & 30000 & 44 & 4 & 7 & 14 & 21 & 30 & 3 & 150 & 463 & 720 & 1063 & 81\\
HTRU\_2 & 17898 & 57 & 2 & 3 & 6 & 8 & 12 & 4 & 100 & 209 & 310 & 434 & 72\\
Ionosphere & 351 & 143 & 2 & 3 & 4 & 5 & 6 & 15 & 147 & 181 & 213 & 254 & 688\\
appendicitis & 106 & 530 & 2 & 7 & 8 & 9 & 9 & 22 & 421 & 447 & 479 & 512 & 1497\\
hepatitis & 155 & 361 & 2 & 12 & 13 & 14 & 14 & 41 & 850 & 916 & 956 & 1075 & 3360
\end{tabular}
}
\end{table}

%% file: comparison_vs_dl85.tex
\begin{table}
\caption{Comparison of DL8.5 (\cite{dl85}) and our approach, MurTree. For each dataset, the number of instances ($\mathcal{D}$), number of binary features ($\mathcal{F}$), and number of classes ($\mathcal{C}$) are displayed. Entries display runtime in seconds (rounded to nearest integer) to compute the optimal classification tree of depth four and five. Time limit set to ten minutes. Timeouts denoted as --- \label{table:vsdl85}}%
\centering
\resizebox{0.7\columnwidth}{!}{
\def\d{@{\hspace*{1.7mm}}}
\begin{tabular}{ \d l | r | r | r | r  r | r  r \d }

	Name & $\mathcal{D}$ & $\mathcal{F}$ & $\mathcal{C}$ & \multicolumn{2}{c | }{Depth=4} & \multicolumn{2}{c}{Depth=5} \\ 
	  &   &   &   & DL8.5 & MurTree & DL8.5 & MurTree \\ \hline
anneal & 812 & 93 & 2 & 55 & $\bm{<}\textbf{0.5}$& --- & \textbf{4} \\
audiology & 216 & 148 & 2 & 99 & $\bm{<}\textbf{0.5}$& $\bm{<}\textbf{0.5}$& $\bm{<}\textbf{0.5}$ \\
australian-credit & 653 & 125 & 2 & 383 & \textbf{2}& --- & \textbf{46} \\
breast-wisconsin & 683 & 120 & 2 & 188 & \textbf{1}& --- & \textbf{2} \\
diabetes & 768 & 112 & 2 & 421 & \textbf{2}& --- & \textbf{83} \\
german-credit & 1000 & 112 & 2 & 326 & \textbf{2}& --- & \textbf{86} \\
heart-cleveland & 296 & 95 & 2 & 108 & $\bm{<}\textbf{0.5}$& --- & \textbf{7} \\
hepatitis & 137 & 68 & 2 & 13 & $\bm{<}\textbf{0.5}$& 35 & $\bm{<}\textbf{0.5}$ \\
hypothyroid & 3247 & 88 & 2 & 104 & \textbf{2}& --- & \textbf{38} \\
ionosphere & 351 & 445 & 2 & --- & \textbf{89}& --- & \textbf{194} \\
kr-vs-kp & 3196 & 73 & 2 & 49 & \textbf{1}& --- & \textbf{16} \\
letter & 20000 & 224 & 2 & --- & \textbf{296}& ---& --- \\
lymph & 148 & 68 & 2 & 8 & $\bm{<}\textbf{0.5}$& 7 & $\bm{<}\textbf{0.5}$ \\
mushroom & 8124 & 119 & 2 & 26 & \textbf{1}& 23 & $\bm{<}\textbf{0.5}$ \\
pendigits & 7494 & 216 & 2 & --- & \textbf{76}& --- & \textbf{464} \\
primary-tumor & 336 & 31 & 2 & 1 & $\bm{<}\textbf{0.5}$& 11 & $\bm{<}\textbf{0.5}$ \\
segment & 2310 & 235 & 2 & 1 & $\bm{<}\textbf{0.5}$& 1 & $\bm{<}\textbf{0.5}$ \\
soybean & 630 & 50 & 2 & 3 & $\bm{<}\textbf{0.5}$& 36 & $\bm{<}\textbf{0.5}$ \\
splice-1 & 3190 & 287 & 2 & --- & \textbf{133}& ---& --- \\
tic-tac-toe & 958 & 27 & 2 & 1 & $\bm{<}\textbf{0.5}$& 7 & $\bm{<}\textbf{0.5}$ \\
vehicle & 846 & 252 & 2 & --- & \textbf{11}& --- & \textbf{277} \\
vote & 435 & 48 & 2 & 4 & $\bm{<}\textbf{0.5}$& 26 & $\bm{<}\textbf{0.5}$ \\
yeast & 1484 & 89 & 2 & 186 & \textbf{2}& --- & \textbf{54} \\
fico-binary & 10459 & 17 & 2 & 1 & $\bm{<}\textbf{0.5}$& 6 & \textbf{1} \\
bank\_conv & 4521 & 26 & 2 & 2 & $\bm{<}\textbf{0.5}$& 15 & \textbf{1} \\
biodeg & 1055 & 81 & 2 & 67 & \textbf{1}& --- & \textbf{22} \\
default\_credit & 30000 & 44 & 4 & 155 & \textbf{4}& --- & \textbf{68} \\
HTRU\_2 & 17898 & 57 & 2 & 64 & \textbf{2}& --- & \textbf{30} \\
Ionosphere & 351 & 143 & 2 & 126 & \textbf{1}& 316 & \textbf{2} \\
magic04 & 19020 & 79 & 2 & 244 & \textbf{4}& --- & \textbf{106} \\
messidor & 1151 & 24 & 2 & $\bm{<}\textbf{0.5}$& $\bm{<}\textbf{0.5}$& 5 & $\bm{<}\textbf{0.5}$ \\
spambase & 4601 & 132 & 2 & --- & \textbf{8}& --- & \textbf{268} \\
Statlog\_satellite & 4435 & 385 & 6 & --- & \textbf{320}& ---& --- \\
Statlog\_shuttle & 43500 & 181 & 7 & --- & \textbf{25}& ---& --- \\
appendicitis & 106 & 530 & 2 & --- & \textbf{7}& --- & \textbf{422} \\
australian & 690 & 1163 & 2 & --- & \textbf{386}& ---& --- \\
backache & 180 & 475 & 2 & --- & \textbf{8}& --- & \textbf{176} \\
cancer & 683 & 89 & 2 & 16 & $\bm{<}\textbf{0.5}$& 301 & \textbf{5} \\
cleve & 303 & 395 & 2 & --- & \textbf{4}& --- & \textbf{500} \\
colic & 368 & 415 & 2 & --- & \textbf{17}& ---& --- \\
haberman & 306 & 92 & 2 & 14 & $\bm{<}\textbf{0.5}$& 293 & \textbf{4} \\
heart-statlog & 270 & 381 & 2 & --- & \textbf{6}& --- & \textbf{383} \\
hepatitis & 155 & 361 & 2 & --- & \textbf{4}& --- & \textbf{119} \\
hungarian & 294 & 330 & 2 & --- & \textbf{4}& --- & \textbf{194} \\
new-throid & 215 & 334 & 3 & --- & \textbf{21}& ---& --- \\
promoters & 106 & 334 & 2 & --- & \textbf{1}& --- & \textbf{1} \\
shuttleM & 14500 & 691 & 2 & --- & \textbf{42}& ---& --- \\
\end{tabular}
}
\end{table}

%% file: figure_out_of_sample_accuracy.tex
\begin{figure}
\centering
\subfloat{
\begin{tikzpicture}[font=\large,scale=.49]
\begin{axis}[
width=\textwidth,
ybar,
bar width=10pt,
xlabel={Difference in accuracy in percentage on the test set},
ylabel={Number of datasets with given difference},
ymin=0,
ytick=\empty,
xtick=data,
axis x line=bottom,
axis y line=left,
enlarge x limits=0.1,
symbolic x coords={-5, -3, -2, -1, 0, 1, 2, 3, 4, 7, 8, 10},
xticklabel style={anchor=base,yshift=-\baselineskip},
nodes near coords={\pgfmathprintnumber\pgfplotspointmeta}
]
\addplot[fill=lightgray] coordinates {(-5, 1) (-3, 5) (-2, 10) (-1, 11) (0, 26) (1, 12) (2, 5) (3, 6) (4, 2) (7, 2) (8, 1) (10, 1)};
\end{axis}
\end{tikzpicture}
}\hfill
\subfloat{
\begin{tikzpicture}[font=\large,scale=.49]
\begin{axis}[
width=\textwidth,
ybar,
bar width=10pt,
xlabel={Runtime ranges in seconds},
ylabel={Number of datasets with given runtime},
ymin=0,
ytick=\empty,
xtick=data,
axis x line=bottom,
axis y line=left,
enlarge x limits=0.1,
symbolic x coords={0-4, 5-9, 10-59, 180-299, 300-1799, 1800+},
xticklabel style={anchor=base,yshift=-\baselineskip},
nodes near coords={\pgfmathprintnumber\pgfplotspointmeta}
]
\addplot[fill=lightgray] coordinates {(0-4, 71) (5-9, 2) (10-59, 6) (180-299, 1) (300-1799, 3) (1800+, 1)};
\end{axis}
\end{tikzpicture}
}\caption*{MT-F: Depth and number of nodes fixed to the values of the CART tree}

\subfloat{
\begin{tikzpicture}[font=\large,scale=.49]
\begin{axis}[
width=\textwidth,
ybar,
bar width=10pt,
xlabel={Difference in accuracy in percentage on the test set},
ylabel={Number of datasets with given difference},
ymin=0,
ytick=\empty,
xtick=data,
axis x line=bottom,
axis y line=left,
enlarge x limits=0.1,
symbolic x coords={-2, -1, 0, 1, 2, 3, 4, 6, 7, 8, 10},
xticklabel style={anchor=base,yshift=-\baselineskip},
nodes near coords={\pgfmathprintnumber\pgfplotspointmeta}
]
\addplot[fill=lightgray] coordinates {(-2, 1) (-1, 9) (0, 32) (1, 14) (2, 12) (3, 6) (4, 3) (6, 1) (7, 2) (8, 1) (10, 1)};
\end{axis}
\end{tikzpicture}
}\hfill
\subfloat{
\begin{tikzpicture}[font=\large,scale=.49]
\begin{axis}[
width=\textwidth,
ybar,
bar width=10pt,
xlabel={Runtime ranges in seconds},
ylabel={Number of datasets with given runtime},
ymin=0,
ytick=\empty,
xtick=data,
axis x line=bottom,
axis y line=left,
enlarge x limits=0.1,
symbolic x coords={0-4, 5-9, 10-59, 60-179, 300-1799, 1800+},
xticklabel style={anchor=base,yshift=-\baselineskip},
nodes near coords={\pgfmathprintnumber\pgfplotspointmeta}
]
\addplot[fill=lightgray] coordinates {(0-4, 70) (5-9, 3) (10-59, 4) (60-179, 2) (300-1799, 4) (1800+, 1)};
\end{axis}
\end{tikzpicture}
}\caption*{MT-R: The CART tree provides the upper bound on the depth and number of nodes}

\subfloat{
\begin{tikzpicture}[font=\large,scale=.49]
\begin{axis}[
width=\textwidth,
ybar,
bar width=10pt,
xlabel={Difference in accuracy in percentage on the test set},
ylabel={Number of datasets with given difference},
ymin=0,
ytick=\empty,
xtick=data,
axis x line=bottom,
axis y line=left,
enlarge x limits=0.1,
symbolic x coords={-1, 0, 1, 2, 3, 4, 5, 6, 7, 8, 19},
xticklabel style={anchor=base,yshift=-\baselineskip},
nodes near coords={\pgfmathprintnumber\pgfplotspointmeta}
]
\addplot[fill=lightgray] coordinates {(-1, 4) (0, 27) (1, 14) (2, 17) (3, 7) (4, 6) (5, 1) (6, 1) (7, 3) (8, 1) (19, 1)};
\end{axis}
\end{tikzpicture}
}\hfill
\subfloat{
\begin{tikzpicture}[font=\large,scale=.49]
\begin{axis}[
width=\textwidth,
ybar,
bar width=10pt,
xlabel={Runtime ranges in seconds},
ylabel={Number of datasets with given runtime},
ymin=0,
ytick=\empty,
xtick=data,
axis x line=bottom,
axis y line=left,
enlarge x limits=0.1,
symbolic x coords={0-4, 5-9, 10-59, 60-179, 300-1799, 1800+},
xticklabel style={anchor=base,yshift=-\baselineskip},
nodes near coords={\pgfmathprintnumber\pgfplotspointmeta}
]
\addplot[fill=lightgray] coordinates {(0-4, 53) (5-9, 6) (10-59, 12) (60-179, 6) (300-1799, 5) (1800+, 2)};
\end{axis}
\end{tikzpicture}
}\caption*{MT-A: full tuning, i.e.,$depth \in \{1, 2, 3, 4\}$ and $num\_nodes \in \{1, 2, ..., 15\}$}

\caption{Performance comparison of our MurTree approach against CART on 84 datasets using difference tuning strategies for the depth and number of nodes of the optimal tree.}
\label{figure:outofsample}
\end{figure}

%% file: figure_out_of_sample_vs_forest.tex
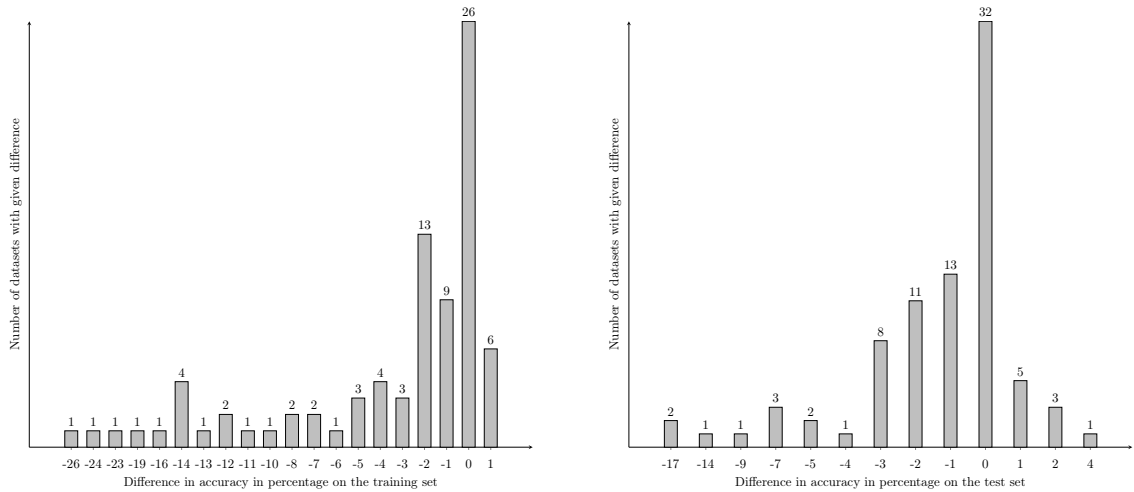
\begin{figure}
\centering
\subfloat{
\begin{tikzpicture}[font=\small,scale=.49]
\begin{axis}[
width=\textwidth,
ybar,
bar width=10pt,
xlabel={Difference in accuracy in percentage on the training set},
ylabel={Number of datasets with given difference},
ymin=0,
ytick=\empty,
xtick=data,
axis x line=bottom,
axis y line=left,
enlarge x limits=0.1,
symbolic x coords={-26, -24, -23, -19, -16, -14, -13, -12, -11, -10, -8, -7, -6, -5, -4, -3, -2, -1, 0, 1},
xticklabel style={anchor=base,yshift=-\baselineskip},
nodes near coords={\pgfmathprintnumber\pgfplotspointmeta}
]
\addplot[fill=lightgray] coordinates {(-26, 1) (-24, 1) (-23, 1) (-19, 1) (-16, 1) (-14, 4) (-13, 1) (-12, 2) (-11, 1) (-10, 1) (-8, 2) (-7, 2) (-6, 1) (-5, 3) (-4, 4) (-3, 3) (-2, 13) (-1, 9) (0, 26) (1, 6)};
\end{axis}
\end{tikzpicture}
}\hfill
\subfloat{
\begin{tikzpicture}[font=\small,scale=.49]
\begin{axis}[
width=\textwidth,
ybar,
bar width=10pt,
xlabel={Difference in accuracy in percentage on the test set},
ylabel={Number of datasets with given difference},
ymin=0,
ytick=\empty,
xtick=data,
axis x line=bottom,
axis y line=left,
enlarge x limits=0.1,
symbolic x coords={-17, -14, -9, -7, -5, -4, -3, -2, -1, 0, 1, 2, 4},
xticklabel style={anchor=base,yshift=-\baselineskip},
nodes near coords={\pgfmathprintnumber\pgfplotspointmeta}
]
\addplot[fill=lightgray] coordinates {(-17, 2) (-14, 1) (-9, 1) (-7, 3) (-5, 2) (-4, 1) (-3, 8) (-2, 11) (-1, 13) (0, 32) (1, 5) (2, 3) (4, 1)};
\end{axis}
\end{tikzpicture}
}\caption{Accuracy comparison of our MurTree (MT-A) approach against random forests on 84 datasets on both the training and test set.}
\label{figure:vsForests}
\end{figure}

%% file: optimal_decision_trees.bbl
\begin{thebibliography}{45}
\providecommand{\natexlab}[1]{#1}
\providecommand{\url}[1]{\texttt{#1}}
\expandafter\ifx\csname urlstyle\endcsname\relax
  \providecommand{\doi}[1]{doi: #1}\else
  \providecommand{\doi}{doi: \begingroup \urlstyle{rm}\Url}\fi

\bibitem[Aghaei et~al.(2019)Aghaei, Azizi, and Vayanos]{aghaei2019learning}
Sina Aghaei, Mohammad~Javad Azizi, and Phebe Vayanos.
\newblock Learning optimal and fair decision trees for non-discriminative
  decision-making.
\newblock In \emph{Proceedings of AAAI}, 2019.

\bibitem[Aghaei et~al.(2020)Aghaei, Gomez, and Vayanos]{aghaei2020learning}
Sina Aghaei, Andres Gomez, and Phebe Vayanos.
\newblock Learning optimal classification trees: Strong max-flow formulations.
\newblock \emph{arXiv preprint arXiv:2002.09142}, 2020.

\bibitem[Aglin et~al.(2020{\natexlab{a}})Aglin, Nijssen, and Schaus]{dl85}
Ga{\"e}l Aglin, Siegfried Nijssen, and Pierre Schaus.
\newblock Learning optimal decision trees using caching branch-and-bound
  search.
\newblock In \emph{Proceedings of AAAI}, 2020{\natexlab{a}}.

\bibitem[Aglin et~al.(2020{\natexlab{b}})Aglin, Nijssen, Schaus, and
  ICTEAM]{aglin2020pydl8}
Ga{\"e}l Aglin, Siegfried Nijssen, Pierre Schaus, and UCLouvain ICTEAM.
\newblock Pydl8. 5: a library for learning optimal decision trees.
\newblock In \emph{Proceedings of IJCAI}, 2020{\natexlab{b}}.

\bibitem[Angelino et~al.(2017)Angelino, Larus-Stone, Alabi, Seltzer, and
  Rudin]{angelino2017learning}
Elaine Angelino, Nicholas Larus-Stone, Daniel Alabi, Margo Seltzer, and Cynthia
  Rudin.
\newblock Learning certifiably optimal rule lists for categorical data.
\newblock \emph{The Journal of Machine Learning Research}, 2017.

\bibitem[Ashok et~al.(2020)Ashok, Jackermeier, Jagtap,
  K{\v{r}}et{\'\i}nsk{\`y}, Weininger, and Zamani]{ashok2020dtcontrol}
Pranav Ashok, Mathias Jackermeier, Pushpak Jagtap, Jan
  K{\v{r}}et{\'\i}nsk{\`y}, Maximilian Weininger, and Majid Zamani.
\newblock dtcontrol: decision tree learning algorithms for controller
  representation.
\newblock In \emph{Proceedings of the International Conference on Hybrid
  Systems: Computation and Control}, 2020.

\bibitem[Avellaneda(2020)]{avellanedaefficient}
Florent Avellaneda.
\newblock Efficient inference of optimal decision trees.
\newblock In \emph{Proceedings of AAAI}, 2020.

\bibitem[Bastani et~al.(2018)Bastani, Pu, and
  Solar-Lezama]{bastani2018verifiable}
Osbert Bastani, Yewen Pu, and Armando Solar-Lezama.
\newblock Verifiable reinforcement learning via policy extraction.
\newblock In \emph{Proceedings of NeurIPS}, 2018.

\bibitem[Bertsimas and Dunn(2017)]{bertsimas2017optimal}
Dimitris Bertsimas and Jack Dunn.
\newblock Optimal classification trees.
\newblock \emph{Machine Learning}, 106\penalty0 (7), 2017.

\bibitem[Bertsimas and Shioda(2007)]{bertsimas2007classification}
Dimitris Bertsimas and Romy Shioda.
\newblock Classification and regression via integer optimization.
\newblock \emph{Operations Research}, 55\penalty0 (2), 2007.

\bibitem[Bessiere et~al.(2009)Bessiere, Hebrard, and O'Sullivan]{firstcp}
Christian Bessiere, Emmanuel Hebrard, and Barry O'Sullivan.
\newblock Minimising decision tree size as combinatorial optimisation.
\newblock In \emph{Proceedings of {CP}}, 2009.

\bibitem[Blanc et~al.(2020)Blanc, Lange, and Tan]{blanc2020provable}
Guy Blanc, Jane Lange, and Li-Yang Tan.
\newblock Provable guarantees for decision tree induction: the agnostic
  setting.
\newblock \emph{Proceedings of ICML}, 2020.

\bibitem[Blanquero et~al.(2020)Blanquero, Carrizosa, Molero-R{\'\i}o, and
  Morales]{blanquero2020sparsity}
Rafael Blanquero, Emilio Carrizosa, Cristina Molero-R{\'\i}o, and
  Dolores~Romero Morales.
\newblock Sparsity in optimal randomized classification trees.
\newblock \emph{European Journal of Operational Research}, 2020.

\bibitem[Breiman et~al.(1984)Breiman, Friedman, Olshen, and
  Stone]{breiman1984classification}
Leo Breiman, JH~Friedman, RA~Olshen, and CJ~Stone.
\newblock Classification and regression trees.
\newblock \emph{Cole Statistics/Probability Series}, 1984.

\bibitem[Carrizosa et~al.(2021)Carrizosa, Molero-Río, and
  Romero~Morales]{surveyMathematicalTrees}
Emilio Carrizosa, Cristina Molero-Río, and Dolores Romero~Morales.
\newblock Mathematical optimization in classification and regression trees,
  2021.

\bibitem[Demirovi\'c and Stuckey(2021)]{biObjTrees}
Emir Demirovi\'c and Peter Stuckey.
\newblock Optimal decision trees for nonlinear metrics.
\newblock In \emph{Proceedings of AAAI}, 2021.

\bibitem[Elmachtoub et~al.(2020)Elmachtoub, Liang, and
  McNellis]{elmachtoub2020decision}
Adam~N Elmachtoub, Jason Cheuk~Nam Liang, and Ryan McNellis.
\newblock Decision trees for decision-making under the predict-then-optimize
  framework.
\newblock \emph{Proceedings of ICML}, 2020.

\bibitem[Fayyad and Irani(1993)]{DBLP:conf/ijcai/FayyadI93}
Usama~M. Fayyad and Keki~B. Irani.
\newblock Multi-interval discretization of continuous-valued attributes for
  classification learning.
\newblock In \emph{Proceedings of IJCAI}, 1993.

\bibitem[Garey(1972)]{garey1972optimal}
Michael~R Garey.
\newblock Optimal binary identification procedures.
\newblock \emph{SIAM Journal on Applied Mathematics}, 23\penalty0 (2), 1972.

\bibitem[Hehn et~al.(2019)Hehn, Kooij, and Hamprecht]{hehn2019end}
Thomas~M Hehn, Julian~FP Kooij, and Fred~A Hamprecht.
\newblock End-to-end learning of decision trees and forests.
\newblock \emph{International Journal of Computer Vision}, 2019.

\bibitem[Hu et~al.(2020)Hu, Siala, Hebrard, and Huguet]{sataccuracy}
Hao Hu, Mohamed Siala, Emmanuel Hebrard, and Marie{-}Jos{\'{e}} Huguet.
\newblock Learning optimal decision trees with maxsat and its integration in
  adaboost.
\newblock In \emph{Proceedings of {IJCAI}}, 2020.

\bibitem[Hu et~al.(2019)Hu, Rudin, and Seltzer]{hu2019optimal}
Xiyang Hu, Cynthia Rudin, and Margo Seltzer.
\newblock Optimal sparse decision trees.
\newblock In \emph{Proceedings of NeurIPS}, 2019.

\bibitem[Hyafil and Rivest(1976)]{NPhardTrees}
Laurent Hyafil and Ronald~L Rivest.
\newblock Constructing optimal binary decision trees is {NP}-complete.
\newblock \emph{Information Processing Letters}, 5\penalty0 (1), 1976.

\bibitem[Janota and Morgado(2020)]{janota2020sat}
Mikol{\'a}{\v{s}} Janota and Ant{\'o}nio Morgado.
\newblock Sat-based encodings for optimal decision trees with explicit paths.
\newblock In \emph{Proceedings of SAT}, 2020.

\bibitem[Kamiran et~al.(2010)Kamiran, Calders, and
  Pechenizkiy]{kamiran2010discrimination}
Faisal Kamiran, Toon Calders, and Mykola Pechenizkiy.
\newblock Discrimination aware decision tree learning.
\newblock In \emph{IEEE International Conference on Data Mining}, 2010.

\bibitem[Kim(2015)]{kim2015package}
HyunJi Kim.
\newblock Package ‘discretization’ in cran-r.
\newblock \url{https://CRAN.R-project.org/package=discretization}, 2015.
\newblock [Online; accessed 21-May-2020].

\bibitem[Kontschieder et~al.(2015)Kontschieder, Fiterau, Criminisi, and
  Rota~Bulo]{kontschieder2015deep}
Peter Kontschieder, Madalina Fiterau, Antonio Criminisi, and Samuel Rota~Bulo.
\newblock Deep neural decision forests.
\newblock In \emph{Proceedings of the IEEE International Conference on Computer
  Vision}, 2015.

\bibitem[Lin et~al.(2020)Lin, Zhong, Hu, Rudin, and Seltzer]{sparseICML}
Jimmy Lin, Chudi Zhong, Diane Hu, Cynthia Rudin, and Margo Seltzer.
\newblock Generalized and scalable optimal sparse decision trees.
\newblock In \emph{Proceedings of ICML}, 2020.

\bibitem[Narodytska et~al.(2018)Narodytska, Ignatiev, Pereira, and
  Marques-Silva]{narodytska2018learning}
Nina Narodytska, Alexey Ignatiev, Filipe Pereira, and Joao Marques-Silva.
\newblock Learning optimal decision trees with {SAT}.
\newblock In \emph{Proceedings of IJCAI}, 2018.

\bibitem[Nijssen and Fromont(2007)]{dl8}
Siegfried Nijssen and Elisa Fromont.
\newblock Mining optimal decision trees from itemset lattices.
\newblock In \emph{Proceedings of SIGKDD}, 2007.

\bibitem[Nijssen and Fromont(2010)]{nijssen2010optimal}
Siegfried Nijssen and Elisa Fromont.
\newblock Optimal constraint-based decision tree induction from itemset
  lattices.
\newblock \emph{Data Mining and Knowledge Discovery}, 21\penalty0 (1), 2010.

\bibitem[Ordyniak and Szeider(2021)]{parametrizedTrees}
Sebastian Ordyniak and Stefan Szeider.
\newblock Parameterized complexity of small decision tree learning.
\newblock In \emph{Proceedings of AAAI}, 2021.

\bibitem[Pedregosa et~al.(2011)Pedregosa, Varoquaux, Gramfort, Michel, Thirion,
  Grisel, Blondel, Prettenhofer, Weiss, Dubourg, Vanderplas, Passos,
  Cournapeau, Brucher, Perrot, and Duchesnay]{scikit-learn}
F.~Pedregosa, G.~Varoquaux, A.~Gramfort, V.~Michel, B.~Thirion, O.~Grisel,
  M.~Blondel, P.~Prettenhofer, R.~Weiss, V.~Dubourg, J.~Vanderplas, A.~Passos,
  D.~Cournapeau, M.~Brucher, M.~Perrot, and E.~Duchesnay.
\newblock Scikit-learn: Machine learning in {P}ython.
\newblock \emph{Journal of Machine Learning Research}, 12, 2011.

\bibitem[Quinlan(1993)]{c4-5}
Ross Quinlan.
\newblock \emph{C4.5: Programs for Machine Learning}.
\newblock Kaufmann, 1993.

\bibitem[Schidler and Szeider(2021)]{LNSSAT}
Andr\'e Schidler and Stefan Szeider.
\newblock Sat-based decision tree learning for large data sets.
\newblock In \emph{Proceedings of AAAI}, 2021.

\bibitem[Tanno et~al.(2019)Tanno, Arulkumaran, Alexander, Criminisi, and
  Nori]{tanno2019adaptive}
Ryutaro Tanno, Kai Arulkumaran, Daniel Alexander, Antonio Criminisi, and Aditya
  Nori.
\newblock Adaptive neural trees.
\newblock In \emph{Proceedings of ICML}, 2019.

\bibitem[Verhaeghe et~al.(2019)Verhaeghe, Nijssen, Pesant, Quimper, and
  Schaus]{verhaeghe2019learning}
H{\'e}lene Verhaeghe, Siegfried Nijssen, Gilles Pesant, Claude-Guy Quimper, and
  Pierre Schaus.
\newblock Learning optimal decision trees using constraint programming.
\newblock In \emph{Proceedings of CP}, 2019.

\bibitem[Verwer and Zhang(2017)]{verwer2017learning}
Sicco Verwer and Yingqian Zhang.
\newblock Learning decision trees with flexible constraints and objectives
  using integer optimization.
\newblock In \emph{Proceedings of CPAIOR}, 2017.

\bibitem[Verwer and Zhang(2019)]{verwer2019learning}
Sicco Verwer and Yingqian Zhang.
\newblock Learning optimal classification trees using a binary linear program
  formulation.
\newblock In \emph{Proceedings of AAAI}, 2019.

\bibitem[Vidal and Schiffer(2020)]{icmlForest}
Thibaut Vidal and Maximilian Schiffer.
\newblock Born-again tree ensembles.
\newblock In \emph{Proceedings of ICML}, 2020.

\bibitem[Yang et~al.(2019)Yang, Shen, and Gao]{yang2019weighted}
Bin-Bin Yang, Song-Qing Shen, and Wei Gao.
\newblock Weighted oblique decision trees.
\newblock In \emph{Proceedings of the AAAI}, 2019.

\bibitem[Zaki and Gouda(2003)]{diffsets}
Mohammed~J Zaki and Karam Gouda.
\newblock Fast vertical mining using diffsets.
\newblock In \emph{Proceedings of SIGKDD}, 2003.

\bibitem[Zaki(2000)]{zaki2000scalable}
Mohammed~Javeed Zaki.
\newblock Scalable algorithms for association mining.
\newblock \emph{IEEE Transactions on Knowledge and Data Engineering},
  12\penalty0 (3), 2000.

\bibitem[Zhu et~al.(2020)Zhu, Murali, Phan, Nguyen, and
  Kalagnanam]{MIPneurips2020}
Haoran Zhu, Pavankumar Murali, Dzung~T Phan, Lam~M Nguyen, and Jayant~R
  Kalagnanam.
\newblock A scalable mip-based method for learning optimal multivariate
  decision trees.
\newblock In \emph{Proceedings of NeurIPS}, 2020.

\bibitem[Zimmermann and De~Raedt(2009)]{zimmermann2009cluster}
Albrecht Zimmermann and Luc De~Raedt.
\newblock Cluster-grouping: from subgroup discovery to clustering.
\newblock \emph{Machine Learning}, 77\penalty0 (1), 2009.

\end{thebibliography}
